\documentclass[review]{elsarticle}
\usepackage[dvipsnames]{xcolor}
\usepackage{graphicx}
\usepackage[caption=false]{subfig}
\usepackage{multirow}
\usepackage{amsmath,amssymb}
\usepackage{array}
\journal{Elsevier}

\begin{document}

\begin{frontmatter}

\title{Effective Action Recognition with \\Embedded Key Point Shifts}

\author[1]{Haozhi Cao\fnref{fn1}}
\ead{haozhi001@e.ntu.edu.sg}
\author[1]{Yuecong Xu\corref{cor1}\fnref{fn1}}
\ead{xuyu0014@e.ntu.edu.sg}
\author[1]{Jianfei Yang\fnref{fn1}}
\ead{yang0478@e.ntu.edu.sg}
\author[1]{Kezhi Mao\fnref{fn1}}
\ead{ekzmao@ntu.edu.sg}
\author[2]{Jianxiong Yin\fnref{fn3}}
\ead{jianxiongy@nvidia.com}
\author[2]{Simon See\fnref{fn3}}
\ead{ssee@nvidia.com}
\cortext[cor1]{Corresponding author}
\fntext[fn1]{School of Electrical and Electronic Engineering, Nanyang Technological University.}
\fntext[fn2]{NVIDIA AI Tech Centre.}
\address[1]{50 Nanyang Avenue, 639798, Singapore}
\address[2]{3 International Business Park Rd, \#01-20A Nordic European Centre, 609927, Singapore}

\begin{abstract}
Temporal feature extraction is an essential technique in video-based action recognition. Key points have been utilized in skeleton-based action recognition methods but they require costly key point annotation. In this paper, we propose a novel temporal feature extraction module, named Key Point Shifts Embedding Module ($KPSEM$), to adaptively extract channel-wise key point shifts across video frames without key point annotation for temporal feature extraction. Key points are adaptively extracted as feature points with maximum feature values at split regions, while key point shifts are the spatial displacements of corresponding key points. The key point shifts are encoded as the overall temporal features via linear embedding layers in a multi-set manner. Our method achieves competitive performance through embedding key point shifts with trivial computational cost, achieving the state-of-the-art performance of $82.05\%$ on Mini-Kinetics and competitive performance on UCF101, Something-Something-v1 and HMDB51 datasets.
\end{abstract}

\begin{keyword}
Action Recognition\sep Temporal Feature\sep Key Point Shifts
\end{keyword}

\end{frontmatter}

\section{Introduction}
\label{section:intro}
\begin{figure}[!t]
  \centering
  \subfloat[\label{intro:intuition-0}]{
  \includegraphics[width=.45\textwidth]{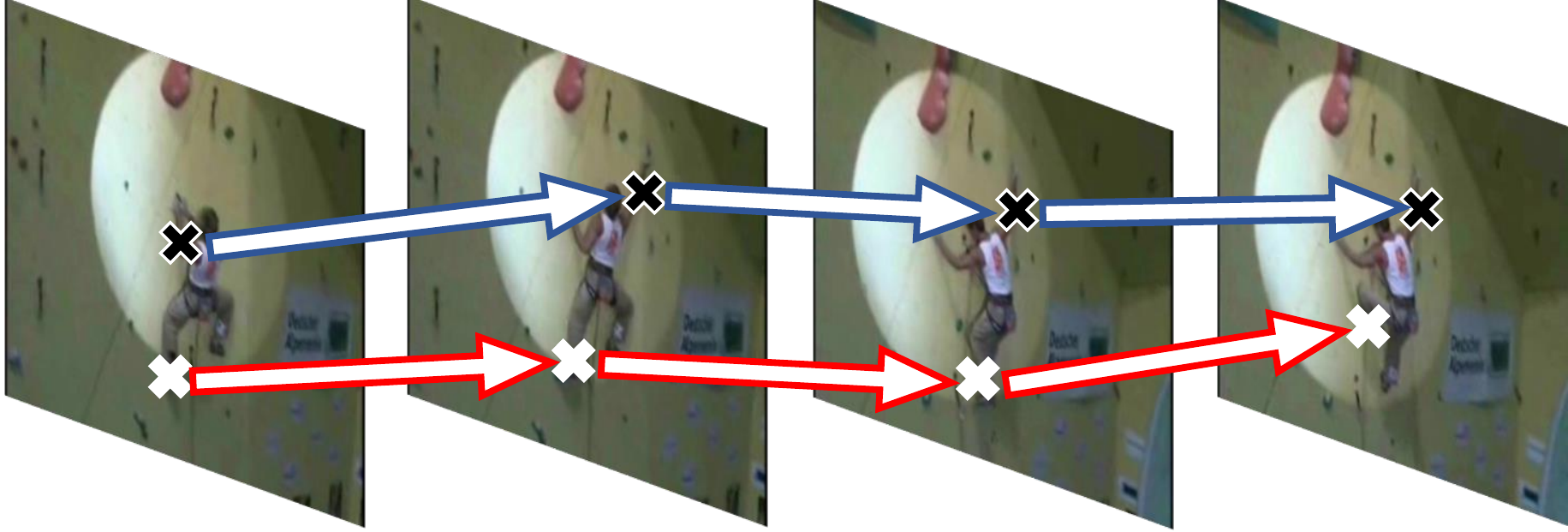}}
  \hfill
  \subfloat[\label{intro:intuition-1}]{
  \includegraphics[width=.45\textwidth]{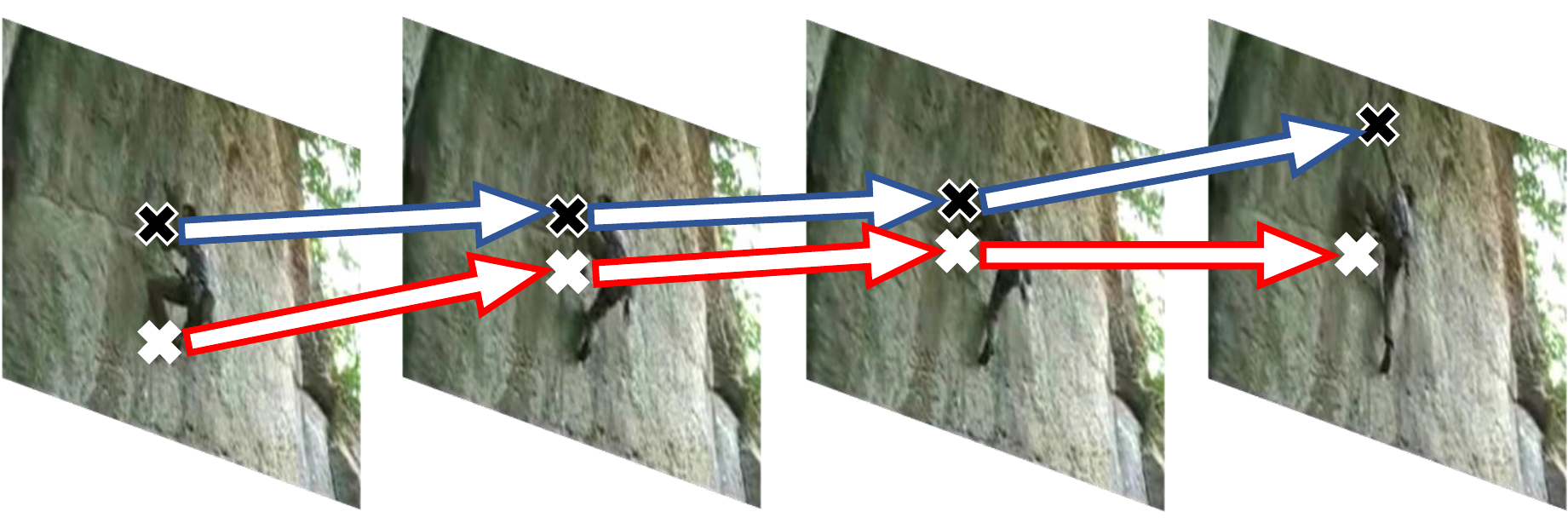}}
  \hfill
  \subfloat[\label{intro:intuition-2}]{
  \includegraphics[width=.45\textwidth]{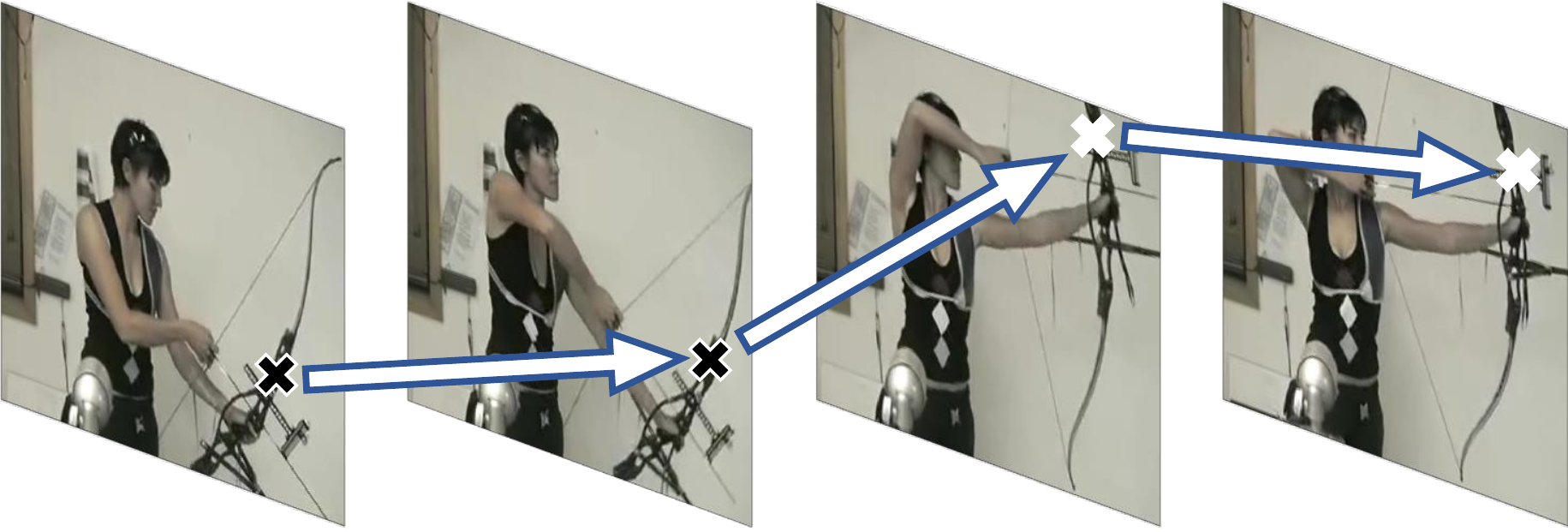}}\\
  \caption{Illustration of using key points and their shifts for action recognition. The key point shifts in Figure~\ref{intro:intuition-0} and Figure~\ref{intro:intuition-1} imply the action ``Climbing" while the key point shifts on Figure~\ref{intro:intuition-2} imply the action ``Archery". Figure~\ref{intro:intuition-0} and Figure~\ref{intro:intuition-1} show that key points and their shifts could be distributed in different locations and the distributions could be different at each frame. Figure~\ref{intro:intuition-2} illustrates an example of the cross-region key point shifts. Crosses and arrows in different colours indicate key points in different regions and different key point shifts. The pictures are best viewed in colour and zoomed in.}
  \label{intro:intuition}
\end{figure}

Action recognition has attracted interest in vision and machine learning communities \cite{dang2020sensor,presti20163d} thanks to its applications such as surveillance \cite{xiang2008activity} and smart homes \cite{yang2018device}. Besides the spatial information in each video frame, videos contain additional temporal information with different properties. Thus, effective modelling of temporal information in videos is critical for accurate action recognition. Intuitively, we humans could effectively recognize the temporal information of action through the key points of actors while ignoring the unrelated environment. Particularly, we focus on the shifts of an actor's key points. For instance, in Figure~\ref{intro:intuition-0} and Figure~\ref{intro:intuition-1}, the actors in different environments are both climbing. We could still recognize the same action for both actors through the movement of the actors' feet and elbows according to key point shifts.

Previous methods used in the skeleton-based action recognition task \cite{li2020learning,si2020skeleton} have proven the effectiveness of using key points and their shifts for recognizing human actions, where the key points and their correspondence across frames are provided as inputs. Skeleton data outlines the key points of actors in the video, suppressing the influence of trivial environment information during the feature extraction process. This increases the robustness of networks when encountering actions with complicated backgrounds. However, additional costs, such as depth sensors \cite{NTURgb} or estimation algorithms \cite{STGCN,PoseEstim}, are needed to annotate the raw video data with human key points for supervision. 

Currently, most methods for action recognition without the annotation of key points would not utilize key point shifts information at all. Instead of learning the action's temporal features, many of these methods may learn more about environment knowledge. This is due to the fact that these methods tend to use all pixels in each frame instead of key points and their shifts, while most of the pixels correspond to the environment information which is not as important as action information. Most of these methods fall into three categories: (1) two-stream CNNs \cite{feichtenhofer2017spatiotemporal,simonyan2014two,wang2016temporal}, (2) 3D CNNs \cite{tran2015learning,tran2018closer,carreira2017quo,hara2018can} and (3) CNNs with learnable feature correlations \cite{wang2018non,wang2018appearance}. Two-stream CNNs model temporal features by inputting pre-computed hand-crafted features, such as optical flow, to a CNN. 3D CNNs extract spatiotemporal features jointly by expanding convolution kernels of 2D CNNs to the temporal dimension but they exhibit inferior performances. More recently, learnable correlations of features across frames \cite{wang2018non,wang2018appearance} are used for temporal modelling. The temporal features are extracted by exploiting the multiplicative interactions between each pixel.

To utilize key point shifts for temporal feature extraction without key point annotation, we propose a novel method to extract and embed key point shifts information of each channel from the high-level feature maps. When high-level spatial features of each frame are extracted through CNN layers, the key points related to the action could be viewed as the points with the maximum feature value of each channel in the high-level feature maps. In addition, in many actions, such key points and their shifts would be distributed in different local regions of the video frames and the distributions of key points would also be slightly different for each frame. For example, for the action ``Climbing" in Figure~\ref{intro:intuition-1}, the shift of the key points corresponding to the elbows and the feet are located at the upper and lower parts, respectively, at the first frame, while both key points are positioned upwards at the last frame. Therefore, to obtain temporal information with respect to the key point shifts at different locations and to cope with the different distributions of key points across all the frames, we propose to split each frame into different regions adaptively with the key points extracted at each region.

Furthermore, for some actions, the key points may not belong to the same region across all the frames. Figure~\ref{intro:intuition-2} depicts such a case where the key point of the bow locates at the lower region at the first two frames while it shifts to the upper region at the last two frames. To extract the cross-region key point shifts correctly, we compute shift weights to indicate the similarity between any pair of key points across adjacent frames. The key point shifts are then calculated as the spatial displacements between the corresponding key points in adjacent frames according to the shift weights. The proposed temporal feature extractor is termed as Key Point Shifts Embedding Module ($KPSEM$), utilizing key point shifts extracted regionally, termed as Regional Key Point Shifts ($RKPS$). Multiple $RKPS$s are obtained through multiple sets of Adaptive Regional Shift Extractor ($AReSE$) under different region separations. The resulting $RKPS$s are encoded through independent embedding layers to constitute more robust temporal features.

In summary, the main contribution of this work is a novel temporal feature extraction module based on key point shifts: Key Point Shifts Embedding Module ($KPSEM$). First, $KPSEM$ is designed to utilize the key point shifts for effective temporal feature extraction without key point annotation. Second, through a multi-set embedding operation, the key point shifts are embedded as effective temporal features of the input video. Third, the extensive experiments on various datasets demonstrate that $KPSEM$ can effectively model temporal features, achieving state-of-the-art performance on the Mini-Kinetics dataset without involving high computational cost.

\section{Related Work}
\label{section:related}
\paragraph{Temporal Feature Extraction}\label{section:TFE} 
To extract temporal features effectively, earlier works \cite{simonyan2014two,feichtenhofer2017spatiotemporal,wang2016temporal} adopt a two-stream strategy where temporal features are extracted in parallel with the spatial features. The temporal features are extracted by feeding a stack of optical flow frames to CNNs. Typical computation methods of optical flow include Lucas–Kanade \cite{lucas1981iterative}, Horn–Schunck \cite{horn1981determining} and TV-L1 \cite{zach2007duality}. More recent two-stream CNNs usually apply TV-L1 \cite{zach2007duality} as the optical flow extraction method due to its robustness and efficiency compared to other methods. Optical flow represents temporal features accurately as it computes pixel-level correlation information across frames. However, the application of optical flow usually forbids end-to-end training of the network, since it requires pre-computation of optical flow before being input to CNNs. Additionally, the process of extracting optical flow is computationally expensive and memory intensive. Therefore, more recent methods try to avoid the need for optical flow.

To address the limitations imposed by utilizing optical flow in two-stream CNNs, later works proposed to extract temporal features jointly with spatial features using 3D CNNs. C3D \cite{tran2015learning}, I3D \cite{carreira2017quo}, P3D \cite{qiu2017p3d} and 3D-ResNet \cite{tran2018closer} all belong to this category. C3D \cite{tran2015learning} is one of the primary works where CNN filters are expanded to the temporal dimension. For faster training, I3D \cite{carreira2017quo} directly inflates 2D CNNs into a 3D structure through endowing filters and pooling kernels with the temporal dimension. Additionally, P3D \cite{qiu2017p3d} reduces the computational cost by simulating a 3D convolution filter with a spatial convolution filter and a separate temporal convolution filter. Subsequent networks, such as 3D-ResNet \cite{tran2018closer}, are deeper and larger 3D CNNs trained on the Kinetics \cite{kay2017kinetics} to retrace the success of deeper 2D CNNs pretrained on ImageNet \cite{deng2009imagenet}. 3D CNNs benefit from end-to-end training and require only RGB input. However, many of these works exhibit inferior results compared to two-stream CNNs. The inferior results could be contributed by the fact that the temporal features are extracted through multiple pooling operations along the temporal dimension. The temporal information which reflects the change in spatial information across time might be lost during the multiple pooling operations. Therefore, 3D CNNs fail to extract effective temporal features, which results in inferior performance. 

To improve the effectiveness of temporal features extracted through CNNs while avoiding the use of optical flow, multiple methods have been proposed. A prominent category is to utilize learnable correlations of features across frames. Inspired by the non-local mean operation for image denoising \cite{buades2005non,li2016novel}, Wang \textit{et al.} \cite{wang2018non} presented the non-local operation to capture correlations on the pixel level as the representation of the temporal features. Similarly, ARTNet \cite{wang2018appearance} is designed such that its relation branch captures the multiplicative interactions between pixels across multiple frames. Recently, the correlation network \cite{wang2020video} utilizes correlation operators to model frame-to-frame correlation in feature maps. The above methods model correlations between all pixels across frames as the temporal features with end-to-end training. However, it is computationally expensive to extract correlations between every single pixel. Moreover, for pixels representing unrelated background information, their correlations might contribute trivial to the overall video features and therefore the correlations computed between these pixels are redundant.

In addition to using learnable feature correlations to extract temporal features, another proposed solution is to enrich the temporal information by extracting temporal features under different frame rates. Specifically, the Slowfast network \cite{feichtenhofer2019slowfast} applies an additional ``Fast" pathway, which is a 3D CNN stream with a higher frame rate to capture temporal features in fine temporal resolution. Similarly, the TPN \cite{yang2020temporal} extracts temporal features by aggregating features extracted under different tempos. These methods aim to extract more temporal information under multiple frame rates. However, they usually include multiple sub-streams for the different frame rates, which might introduce extra network structures and result in computation overhead.

Different from two-stream methods, our work proposed an end-to-end method to extract frame-wise temporal information from pure RGB input. In contrast to previous correlation extraction methods, our proposed method aims to extract the temporal information effectively by extracting key correlations instead of correlations of all pixels or features. Our proposed method utilizes only regional key points and computes their respective shift across successive frames. This implies that our temporal feature extraction method is more computation efficient with less redundant information included. Yet it still brings consistent improvement in temporal feature extraction, supported by stable improvement in action recognition accuracy.

\paragraph{Skeleton-based Action Recognition}\label{section:Skeleton-AR}
Key points and their displacements have been used mainly in skeleton-based action recognition \cite{li2020learning,si2020skeleton}. Most of the skeleton-based methods take skeleton data as the input, which is generated by devices or pose estimation algorithms in the form of 2D or 3D coordinates. These skeleton data already possesses temporal correspondent relationship. The utilization of skeleton-based data excludes the effect of unrelated pixels so that the network can concentrate on the key points and their temporal correlation. Compared to extracting correlations of all pixels or features, the introduction of skeleton data is more effective and efficient in modelling temporal information.

However, such annotations are not available in the first place for most public videos. The generation of skeleton data requires extra cost, such as extra computation resources and additional recording devices. In comparison, we propose a novel method to extract key points in high-level feature maps without the need for collecting and annotating the skeleton data for videos. Empirical results show that our method can bring consistent improvement while resulting in only a trivial amount of extra computational complexity without the need for skeleton data or other key point annotation.

\section{Proposed Method}
\label{section:method}
The ultimate goal of our work is to extract temporal information that can represent the frame-wise movement of spatial features. We believe that key points represent the dominant spatial features, such as parts of human body and objects related to the actions as shown in Figure~\ref{intro:intuition}. Therefore, their shifts can be viewed as effective temporal features. We propose a novel module, Key Point Shifts Extraction Module ($KPSEM$), which utilizes Regional Key Point Shifts ($RKPS$) as a new modality to represent the temporal features of videos. The $RKPS$s are extracted by multiple sets of Adaptive Regional Shift Extractor ($AReSE$) and computed as the relative coordinate displacements between key points. In this section, we first illustrate how the $KPSEM$ is implemented with the CNN backbone as well as its detailed structure. Subsequently, we expound the details of the $AReSE$ with the process of extracting $RKPS$, which is the core of our proposed $KPSEM$. 

\subsection{Overall Structure}\label{section:overall} 
\begin{figure*}[!ht] 
  \centering{ 
  \includegraphics[width=.85\textwidth]{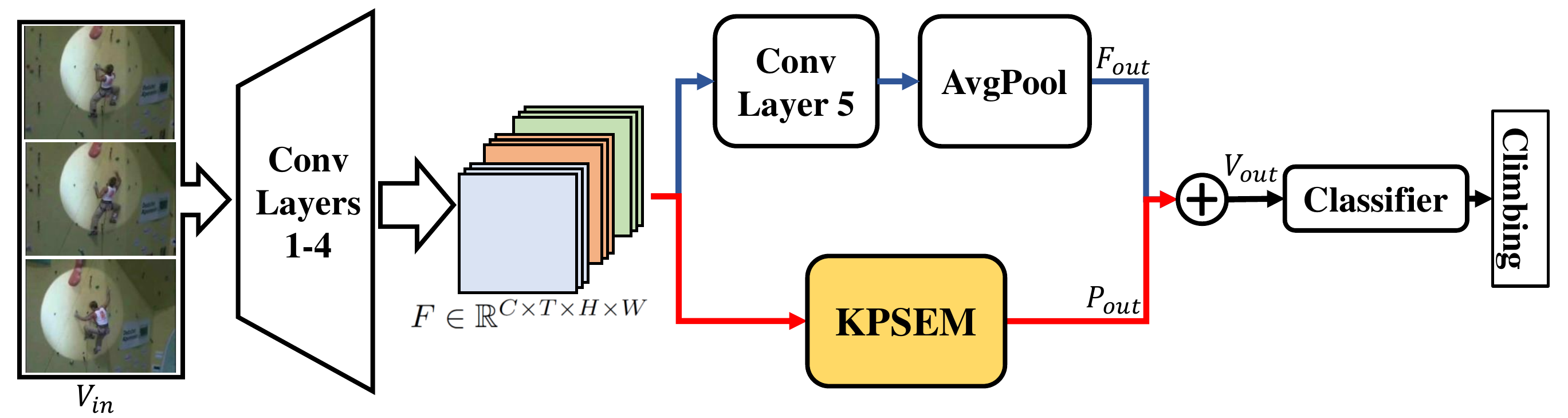}}
  \hfill 
  \caption{The overall structure of our proposed method. $KPSEM$ extracts the temporal features of the video through high-level spatial features $F$ extracted from the CNN backbone. The spatial feature output $F_{out}$ and temporal feature output $P_{out}$ are concatenated and the resulting overall video features $V_{out}$ are passed through a linear classifier. $KPSEM$ can be inserted not only at the location shown in this figure but also after any convolution layer.} 
  \label{method:overall} 
\end{figure*} 

The overall structure of our network is as shown in Figure~\ref{method:overall}. Given an input video $V_{in}$, we utilize CNN to extract its high-level feature maps, denoted as $F\in\mathbb{R}^{C\times T\times H\times W}$ where $C$, $T$, $H$, $W$ are the number of channels, the length of frames, the spatial height and the width of the feature maps, respectively. Our proposed $KPSEM$ extracts temporal features from $F$ by embedding $RKPS$s which are the weighted coordinate differences extracted by $AReSE$. The temporal features from $KPSEM$ concatenate with the spatial features from CNN and the overall video features pass through the linear classifier. The spatial feature output from the CNN is denoted as $F_{out}\in\mathbb{R}^{C_{s}}$ while the temporal feature output of $KPSEM$ is denoted as $P_{out}\in\mathbb{R}^{C_{p}}$. Here $C_{s}$, $C_{p}$ denote the number of channels of the spatial feature output and that of temporal feature output from $KPSEM$, respectively. Subsequently, the overall video features $V_{out}$ are computed by: 
\begin{equation}
    V_{out} = F_{out} \oplus P_{out}, 
\end{equation} 
where $\oplus$ denotes the concatenation operation along the channel dimension. The combined $V_{out} \in\mathbb{R}^{(C_{s} + C_{p})}$ passes through the final classifier. It is worth noting that while the $KPSEM$ is inserted after Conv4 in Fig \ref{method:overall}, the $KPSEM$ is an isolated block which can be inserted at any other location. 

\subsection{Key Point Shifts Embedding Module (KPSEM)}\label{section:KPSEM} 
\begin{figure*}[!ht] 
  \centering{ 
  \includegraphics[width=.85\textwidth]{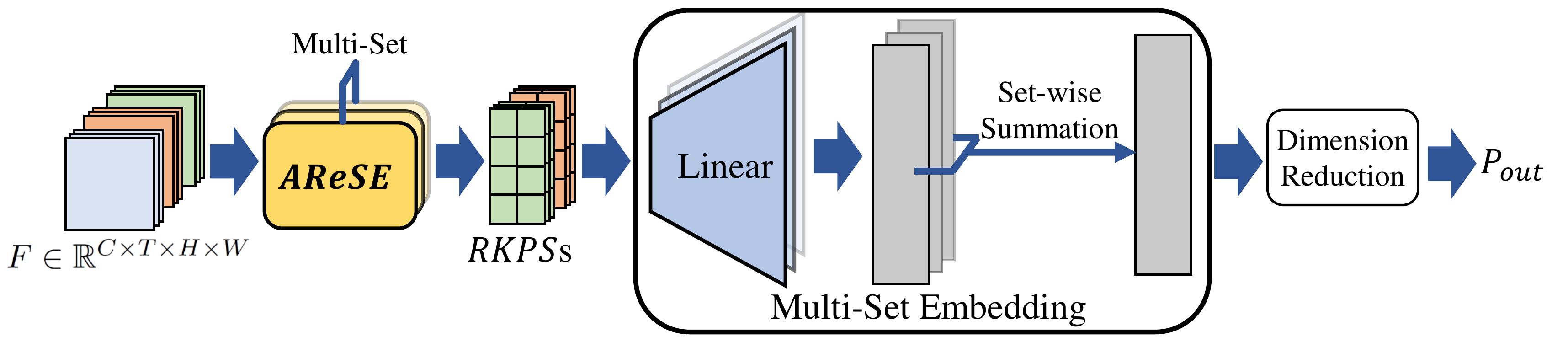}} 
  \hfill 
  \caption{Details of $KPSEM$. From the high-level features $F$ extracted by CNN, $G$ $RKPS$s are extracted by $G$ sets of $AReSE$s. Each $RKPS$ is linearly projected to a higher dimension vector separately. The resulting multi-set embedding is then processed through a set-wise summation to combine all the information from the different sets. The overall temporal feature output is obtained through a final dimension reduction process.} 
  \label{method:S2EM} 
\end{figure*} 

Our proposed $KPSEM$ module extracts the overall temporal features by embedding the Regional Key Point Shifts ($RKPS$) obtained through $AReSE$ (\textit{cf.} Section \ref{section:AReSE}). The $AReSE$ performs an adaptive region separation on high-level feature maps and subsequently extracts the $RKPS$ based on the region separation. Different region separations generate different $RKPS$s. To enhance the robustness of $KPSEM$, we adopt multiple sets of $AReSE$s, each of which performs region separation and $RKPS$ extraction independently. Formally, the $RKPS$s are computed as: 
\begin{equation}\label{equation:multi-setAReSE} 
  RKPS_g = AReSE_g(F), \ \ g = 1,2,...G,
\end{equation} 
where $RKPS_g$ is the $g^{th}$ $RKPS$ extracted by the $g^{th}$ set of $AReSE$, denoted as $AReSE_g$. $G$ is the total number of sets of $AReSE$s. The resulting $RKPS$s are of size ${C}\times {(T-1)}\times {G}\times {2}$. The details about how $AReSE$ extracts $RKPS$ is illustrated in Section \ref{section:AReSE}. 

Subsequently, inspired by the multi-head attention introduced in \cite{vaswani2017attention}, instead of going through a shared linear embedding layer, the $G$ sets of $RKPS$s are linearly projected to dimension $d_e$ separately. This independent embedding operation ensures that our proposed $KPSEM$ can gain more abundant representation from different $RKPS$s. A set-wise summation is then utilized to aggregated information across all the $RKPS$s. More specifically, given $G$ sets of $RKPS$s, the multi-set embedding is computed as: 
\begin{equation} 
  \mathcal{M}(RKPSs) = \sum\limits_{g=1}^{G} \mathcal{L}_g(RKPS_g), 
\end{equation} 
where $\mathcal{M}$ denotes the multi-set embedding operation. $RKPS_g$ is the $g^{th}$ $RKPS$ computed by $AReSE_g$, while $\mathcal{L}_g$ is the $g^{th}$ linear projection for the corresponding $RKPS_g$. The subsequent set-wise summation across all $G$ $RKPS$s directly merges all embeddings. Ultimately, after a dimension reduction procedure, the overall temporal features are obtained. 

We notice that the multi-set mechanism mentioned above plays an important role in the $KPSEM$ module. The utilization of multiple sets of $AReSE$ provides the overall framework with different adaptive regional feature maps. This allows the network to find better local splits with more representative key points. The separated linear embedding layer for each $RKPS$ ensures the abundance of embedded information obtained from the different $RKPS$s. Next, we illustrate how the proposed $AReSE$ conducts adaptive region separation and key point shift extraction in details. 

\subsection{Adaptive Regional Shift Extractor (AReSE)}\label{section:AReSE} 
\begin{figure*}[t] 
  \centering{ 
  \includegraphics[width=.85\textwidth]{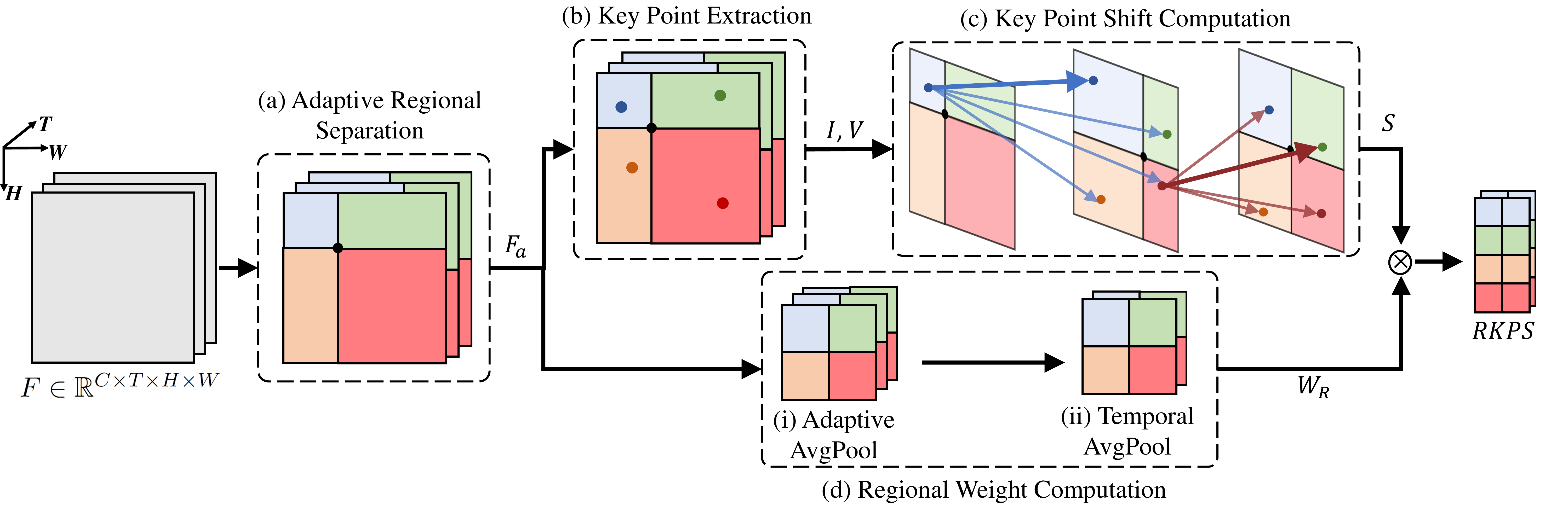}} 
  \hfill 
  \caption{Details of $AReSE$ applied to a single channel. Here each frame is split into $K=4$ regions. The high-level features $F$ extracted by CNN is first separated adaptively, forming the adaptive regional feature maps $F_a$. The key point coordinates $I$ and feature values $V$ are then extracted from each region at each channel. The key point shifts $S$ are computed as the weighted coordinate shift of corresponding regional key points across adjacent frames. The regional weights $W_R$ are generated by performing adaptive average pooling and temporal average pooling operations to indicate the importance of a region. The $RKPS$ is computed as the weighted key point shifts weighted by the regional weights.} 
  \label{method:AReSE} 
\end{figure*}

As mentioned in Section \ref{section:intro}, key point shifts can represent the temporal movement of dominant features of each channel. We propose a novel module named Adaptive Regional Shift Extractor ($AReSE$) to extract key point shifts from high-level feature maps as our temporal features. To cope with the different key point distributions illustrated as Figure~\ref{intro:intuition-1} and preserve local information, we first adaptively separate the feature maps into multiple regions. The key points and their shifts are subsequently extracted from each region to generate Regional Key Point Shifts ($RKPS$). In this section, we illustrate how the proposed $AReSE$ extracts $RKPS$ step by step, including (a) Adaptive Separation of Feature Maps, (b) Key Point Extraction, (c) Key Point Shift Computation and (d) Regional Weight Computation as shown in Figure~\ref{method:AReSE}. 

\paragraph{Adaptive Separation of Feature Maps}\label{subsection:AdaptiveSeparation} 
Taking the high-level features $F$ as the input, the $AReSE$ first separates each frame of feature maps into $K$ regions. Instead of manually setting the boundaries between the regions, the splits of the different regions at each frame are trainable and adaptive in order to adjust to the different key point distributions in each frame. Figure~\ref{method:AReSE} shows an example of adaptive separation of $K=4$ regions, where regions are separated by a single separation centre. For a sequence of $T$ frames, we denote the stack of geometric centres of the feature maps $F$ as $O\in\mathbb{R}^{{T}\times{2}}$, with $O_i$ being the geometric center of the $i^{th}$ frame located at $(x_i,y_i)$. $O$ splits the original feature maps $F$ into $K$ equal regions. The corresponding stack of adaptive separation centres, denoted as $O_a\in\mathbb{R}^{{T}\times{2}}$, is computed by adding an adaptive bias, which is obtained from each frame through a Multilayer Perceptron (MLP). The adaptive separation centres $O_a$ are computed by: 
\begin{eqnarray} 
    B_i & = & \mathcal{I}_i(F_i) = (\Delta x_i, \Delta y_i) \\ 
    O_{a_i} & = & O_i+B_i \label{equation:split-center}, 
\end{eqnarray} 
where $O_i$ and $O_{a_i}$ are the geometric centre and the adaptive separation point of the $i^{th}$ frame $F_i\in\mathbb{R}^{{C}\times{H}\times{W}}$, respectively. $B_i$ is the adaptive bias of the $i^{th}$ frame $F_i$. $\mathcal{I}_i$ is the MLP which generates the adaptive bias $B_i$. The MLPs are trained jointly with the network. Given the adaptive separation centers $O_a$, the high-level features are then separated into $K\times T$ regions, resulting in the adaptive regional feature maps $F_a$ consisting of $\left\{ {F_a}^{(1,1)}, \cdots,{F_a}^{(T,K)}\right\}$ of sizes $\left\{ C\times H^{(1,1)}\times W^{(1,1)}, \cdots,C\times H^{(T,K)}\times W^{(T,K)}\right\}$ as in Figure~\ref{method:AReSE}(a).

The separation of feature maps provides regional key points with local characteristics. If the split of each frame is fixed, different key points that are spatially close to each other may locate within the same region. In such cases, key points except the one with the maximum feature value would be ignored. Since the adaptive separation directly affects the resulting key points, we utilize multi-set of $AReSE$ as mentioned in Section \ref{section:KPSEM} to generate more key points under different region separations. The multi-set operation improves the diversity of feature map separation and therefore improves the robustness of extracted temporal features. 

\paragraph{Key Point Extraction}\label{subsection:KeyPointExtraction} 
Given the high-level feature map of region $r$ located at channel $c$ of frame $f$ denoted as $F_r$, the maximum point is extracted as the key point since the maximum feature value point represents the area of raw pixels with key spatial information. Given ${F_{r}\in\mathbb{R}^{H_r\times W_r}}$, the coordinate as well as the feature value of the key point are extracted as: 
\begin{eqnarray}
  I_{r} & = & \underset{(h,w)}{\arg\max} \ \ {F_{r}}(h,w) = (x_{r}^{max}, y_{r}^{max})\\ 
  V_{r} & = & \max ({F_{r}}) = {F_{r}}(x_{r}^{max}, y_{r}^{max}),
\end{eqnarray} 
where $I_{r}\in\mathbb{R}^{2}$ is the coordinate of the key point at the region $r$ of the feature map. $V_{r}$ is the feature value of the key point.

Note that the key point extraction is operated at each channel in practice. In another word, given a region $F_R\in\mathbb{R}^{C\times H_r\times W_r}$ with multiple channels, the key point extraction generates $C$ key points in total, each of which represents the spatial location of the key feature in its own channel and therefore we can compute the movement of these key features independently in subsequent procedures. Given the adaptive regional feature maps $F_a$ consisting of $\left\{ {F_a}^{(1,1)}, \cdots,{F_a}^{(T,K)}\right\}$, the key point extraction results in key point coordinates $I \in\mathbb{R}^{{K}\times{C}\times{T}\times{2}}$ and key point values $V \in\mathbb{R}^{{K}\times{C}\times{T}}$ indicated as the points with darker color in Figure~\ref{method:AReSE}(b).

\paragraph{Key Point Shift Computation}\label{KPSCompute} 
Given the key point 2D coordinates $I$ and their respective feature values $V$, we compute the regional key point shifts across adjacent frames by two steps, including spatial location difference computation and shift weight computation. 

There are many cases where corresponding key points may not belong to the same region across adjacent frames (e.g. Figure~\ref{intro:intuition-2}). To cope with these situations, we first find the spatial location difference between any pair of regional key points in adjacent frames. The key point shifts are then computed as the weighted sum of the location differences based on the correlation of the key points. This ensures that the computed shift are indeed extracted between two corresponding key points. Given any two key points $I_{i,\alpha}$, $I_{i+1,\beta}$ where $\alpha$, $\beta$ are two regions located at frames $i$, $i+1$, respectively, the spatial location difference of these two key points is computed as: 
\begin{equation}\label{eqn:SLDCompute} 
  \Delta_{i,\alpha,\beta} = I_{i+1,\beta} - I_{i,\alpha}, 
\end{equation} 
where $\Delta_{i,\alpha,\beta}$ is the spatial location difference between key points $I_{i+1,\beta}$ and $I_{i,\alpha}$. In practice, the spatial location computation is operated across any adjacent frames at each channel. Therefore, given the key point coordinates $I\in\mathbb{R}^{{K}\times{C}\times{T}\times{2}}$, the resulting stack of spatial location differences $\Delta$ of size ${K_r\times K_n\times C\times (T-1)\times 2}$ contains all channel-wise spatial location differences between any two key points in adjacent frames. Here we use $K_r$ and $K_n$ to refer the region dimension in the current and the next frame for spatial location differences $\Delta$, respectively, with their values both equal to $K$. 

The key point shifts are then obtained through attending to the spatial location differences of corresponding key points, which should have the strongest correlation. Formally, given the spatial location differences between any two adjacent frames $i$ and $i+1$ denoted as $\Delta_{i}\in\mathbb{R}^{K_r\times K_n \times 2}$, the key point shifts $S_i$ are weighted sums of $\Delta_{i}$ across $K_n$. Here the shift weight $W_{i,\alpha,\beta}$ is related to the correlation between the key points located at region $\alpha$ at the recent frame $i$ and the key point located at region $\beta$ at the next frame $i+1$. The shift weight indicates the probability of the key points at $I_{i,\alpha}$ of the recent frame falling at the location $I_{i+1,\beta}$ at the next frame. Given the stack of spatial location differences $\Delta_i\in\mathbb{R}^{K_r \times K_n \times 2}$, the shift weight and the key point shifts at the region $\alpha$ of the recent frame is computed by: 
\begin{eqnarray} 
  W_{i,\alpha,\beta} & = & \mathcal{G}_n \left( \frac{1}{|V_{i,\alpha}-V_{i+1,\beta}|+0.1} \right ) \\ 
  S_{i,\alpha} & = & \sum_{\beta \in K_n} W_{i,\alpha,\beta} \cdot \Delta_{i,\alpha,\beta}, 
\end{eqnarray} 
where $V_{i,\alpha}$, $V_{i+1,\beta}$ are the feature values of key points in the region $\alpha$ at the recent frame $i$ and the region $\beta$ at the next frame $i+1$ , respectively. The $\mathcal{G}_n$ is the softmax function along the $K_n$ dimension. The above equations show that the correlation between any two key points across adjacent frames is computed by the reciprocal of the difference between their feature values. Similar to the spatial location difference computation mentioned in Equation \ref{eqn:SLDCompute}, the shift weight computation is also applied to all channels and all frames. Given the spatial location differences $\Delta\in\mathbb{R}^{K_r\times K_n\times C\times (T-1)\times 2}$, the resulting key point shifts $S\in\mathbb{R}^{K \times C \times (T-1) \times 2}$ are obtained from the weighted stack of spatial location differences $\Delta$ summed across the $K_n$ dimension per channel. 

\paragraph{Regional Weight Computation}\label{section:RKPSCompute} 
In many videos, the temporal features of the action would be located in a certain region of the video frames. It is therefore reasonable to attend to a certain region while extracting key point shifts. Here an attention mechanism is designed to represent the relative significance of regions across each pair of adjacent frames based on its regional average value. Formally, given the adaptive regional feature maps $F_a$, the regional weight $W_{R}$ is computed by: 
\begin{equation} 
  W_{R} = \mathcal{G}_k(\mathcal{P_T}(\mathcal{P_R}(F_a))). 
\end{equation} 
This implies that the regional weight $W_{R} \in \mathbb{R}^{K \times C \times (T-1)}$ is computed as the average values of $F_a$ over the same region for two adjacent frames. The $\mathcal{P_R}$ is an adaptive average pooling operation along the spatial dimension to generate the mean values of each adaptively split region. Whereas the $\mathcal{P_T}$ is an average pooling operation across adjacent frames with the kernel size set to 2 along the temporal dimension. The $\mathcal{G}_k$ is the softmax function along the $K$ dimension. The computed weight $W_{R}$ represents the attention level to the key point shifts by considering regional feature values across adjacent frames. The resulting $RKPS$ is generated as the weighted key point shifts $S$, weighted by the regional weight $W_{R}$. Formally, given the key point shifts $S$ and the regional weight $W_{R}$, the $RKPS$ is computed as: 
\begin{equation} 
  RKPS = W_{R} \odot S, 
\end{equation} 
where $\odot$ is the element-wise multiplication. The overall $RKPS\in\mathbb{R}^{K\times C\times (T-1)\time 2}$ is therefore the overall weighted key point shifts of corresponding key points across two adjacent frames of the input feature map $F$. 

As mentioned in Section \ref{section:KPSEM}, there are multi-set of $AReSE$s in $KPSEM$, each of which generates $RKPS$ based on its own adaptive region separation. Our experiments as shown in Section \ref{experiment:ablation-visualization} indicate that while the multi-set $AReSE$s can further increase the performance compared to a single $AReSE$, using a single $AReSE$ in $KPSEM$ can also improve the performance compared to the baseline model. 
\section{Experiments}
\label{section:experiment}

In this section, we present our evaluation results of the proposed work. The evaluation is conducted through action recognition experiments on four public benchmark datasets, namely \textbf{Mini-Kinetics} \cite{xie2018rethinking}, \textbf{UCF101} \cite{soomro2012ucf101}, \textbf{Something-Something v1} \cite{Goyal_2017sthsth} and \textbf{HMDB51} \cite{kuehne2011hmdb}. We present state-of-the-art results on Mini-Kinetics dataset, and competitive performances on UCF10, Something-Something v1 and HMDB51 datasets. We also present detailed ablation study performed on HMDB51 \cite{kuehne2011hmdb} dataset to verify our design. We further provide heat maps as well as key point shifts visualization of our proposed framework to justify the effectiveness of our proposed work.

\begin{figure*}[!ht]
  \centering
  \subfloat[\label{exp:settings-1}]{
  \includegraphics[width=.35\textwidth]{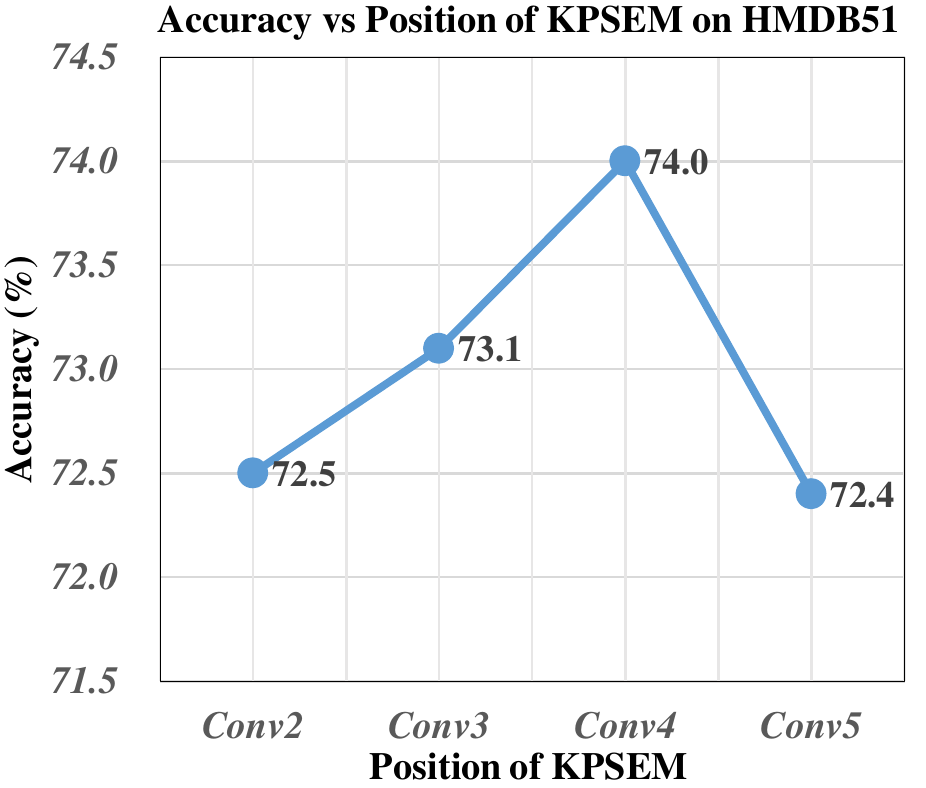}}
  \hspace{.5em}
  \subfloat[\label{exp:settings-2}]{
  \includegraphics[width=.35\textwidth]{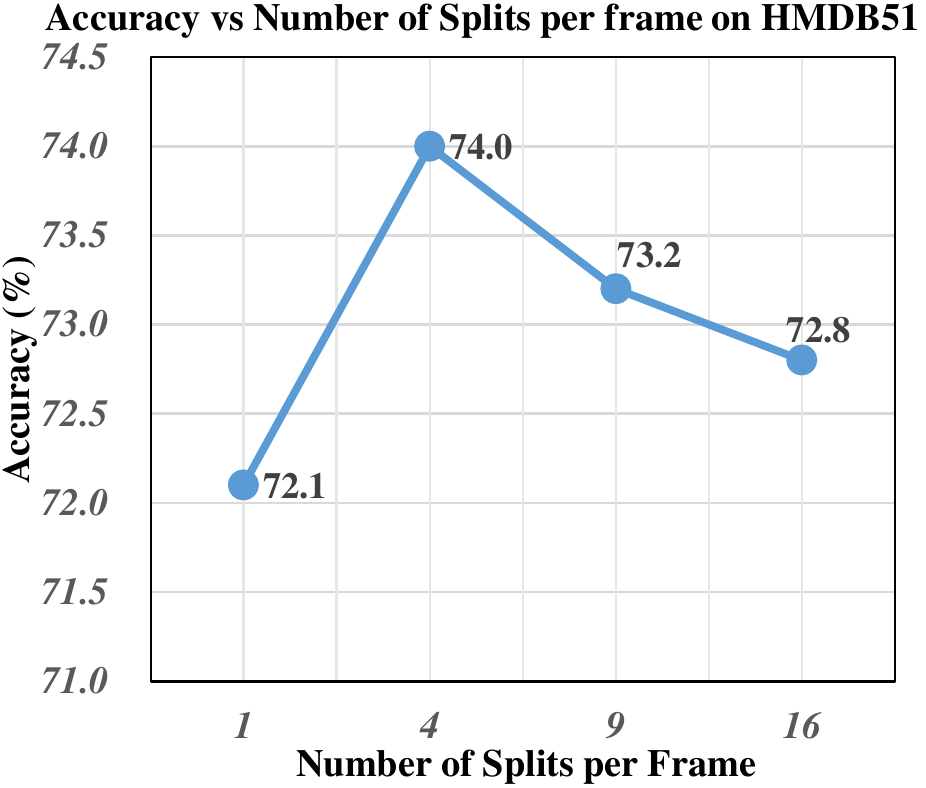}}
  \\[1ex]
  \subfloat[\label{exp:settings-3}]{
  \includegraphics[width=.35\textwidth]{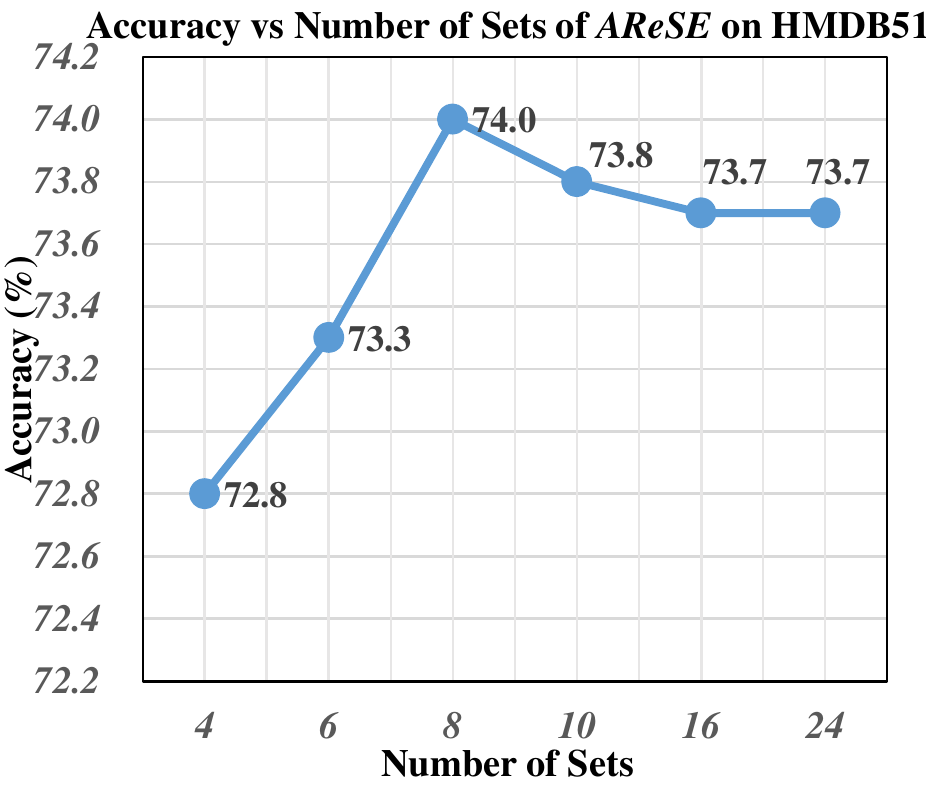}}
  \hspace{.5em}
  \subfloat[\label{exp:settings-4}]{
  \includegraphics[width=.35\textwidth]{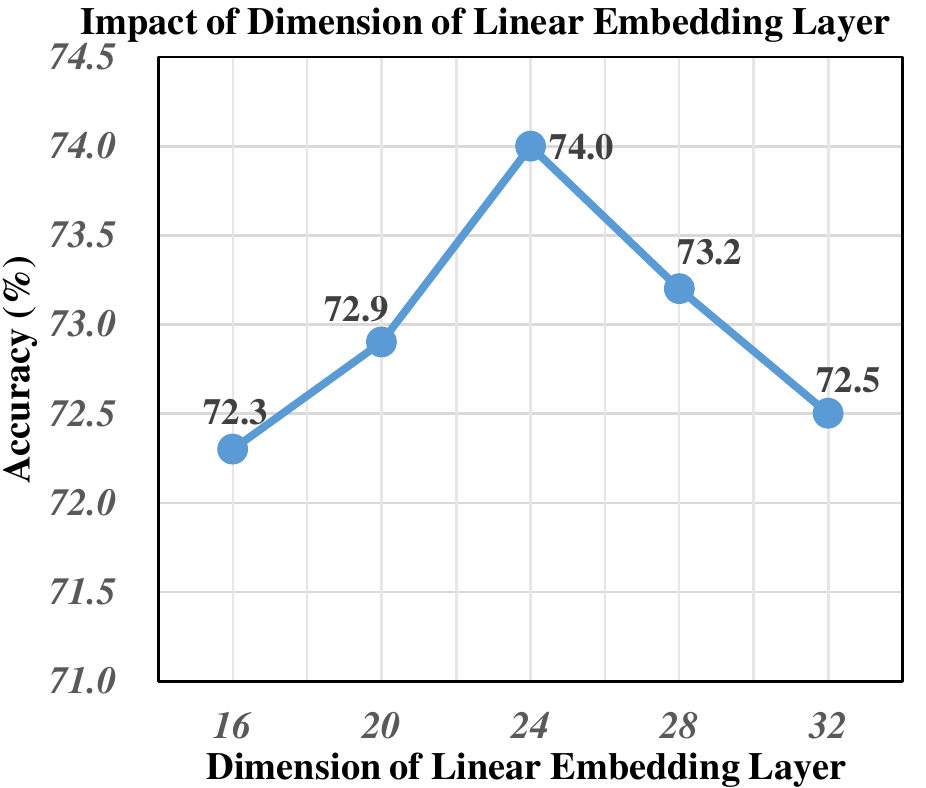}}\\
  \caption{Experiments for best settings of $KPSEM$ with MFNet \cite{chen2018multifiber}. The experiments are conducted on HMDB51 split-1 with a small batch size of 16. (a) Accuracy {\it vs.\ }the position of $KPSEM$ on HMDB51, with $KPSEM$ placed at the end of each Conv layer. (b) Accuracy {\it vs.\ }the number of splits per frame on HMDB51. (c) Accuracy {\it vs.\ }the number of sets of $AReSE$ on HMDB51. (d) Accuracy {\it vs.\ }dimension of the linear embedding layer. Our implementation obtains a 70.8\% accuracy with MFNet, which is much lower than that reported in \cite{chen2018multifiber} mainly due to the much smaller batch size. All settings surpass the baseline by at least $1.3\%$.}
  \label{exp:settings}
\end{figure*}

\subsection{Experimental Settings}\label{experiment:settings}
We conduct experiments on four benchmark datasets of action recognition: Mini-Kinetics, Something-Something v1, UCF101 and HMDB51. \textbf{Mini-Kinetics} is a subset of the Kinetics \cite{kay2017kinetics} dataset, with 200 of its categories. It contains 80K training data and 5K validation data. \textbf{Something-Something v1} \cite{Goyal_2017sthsth} contains 108,499 videos from 174 human-object-interaction action classes, consisted with 86,017 training, 11,522 validation and 10,960 test videos. \textbf{UCF101} \cite{soomro2012ucf101} contains 13,320 videos from 101 action categories. \textbf{HMDB51} \cite{kuehne2011hmdb} contains 51 action categories including a total of 7,000 videos. For UCF101 and HMDB51 datasets, we follow the experiment settings as in \cite{chen2018multifiber,tran2015learning,tran2018closer} that adopt the three train/test splits for evaluation. We report the average top-1 accuracy over the three splits. Our proposed module for temporal feature extraction can be used with any CNN networks. To obtain the state-of-the-art result on Mini-Kinetics and competitive results on UCF101, Something-Something v1 and HMDB51, we instantiate MFNet \cite{chen2018multifiber} thanks to its superior performance on Kinetics. The variant of MFNet combined with $KPSEM$ is referred to as MF-KPSEM.

Our experiments are implemented using PyTorch \cite{paszke2019pytorch}. Following the implementation in \cite{chen2018multifiber}, the input is a frame sequence with each frame of size $224\times224$. Our $KPSEM$ extracts the temporal features from the output of the Conv4 layer of MFNet. We choose to split each frame into $K=4$ splits and adopt $G=8$ sets based on empirical results in Figure~\ref{exp:settings}. The output dimension of the linear embedding layer is set as $24$. This setting exhibits the ultimate results as shown in Figure~\ref{exp:settings}. To accelerate training, we utilize the pretrained model of MFNet \cite{chen2018multifiber} trained on Kinetics \cite{kay2017kinetics}. The stochastic gradient descent algorithm \cite{bottou2010large} is used for optimization, with the weight decay set to 0.0001 and the momentum set to 0.9. Our initial learning rate is set to 0.005. A more detailed settings analysis is illustrated in Section \ref{experiment:ablation-visualization}.

\subsection{Detailed Implementation of $KPSEM$}
\begin{figure*}[!ht]
    \centering
    \subfloat[\label{detail:framework}]{
    \includegraphics[width=.9\linewidth]{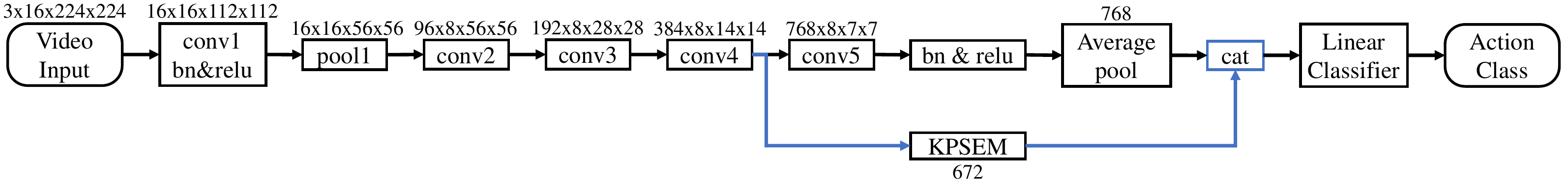}}
    \hfill
    \subfloat[\label{detail:KPSEM}]{
    \includegraphics[width=.9\linewidth]{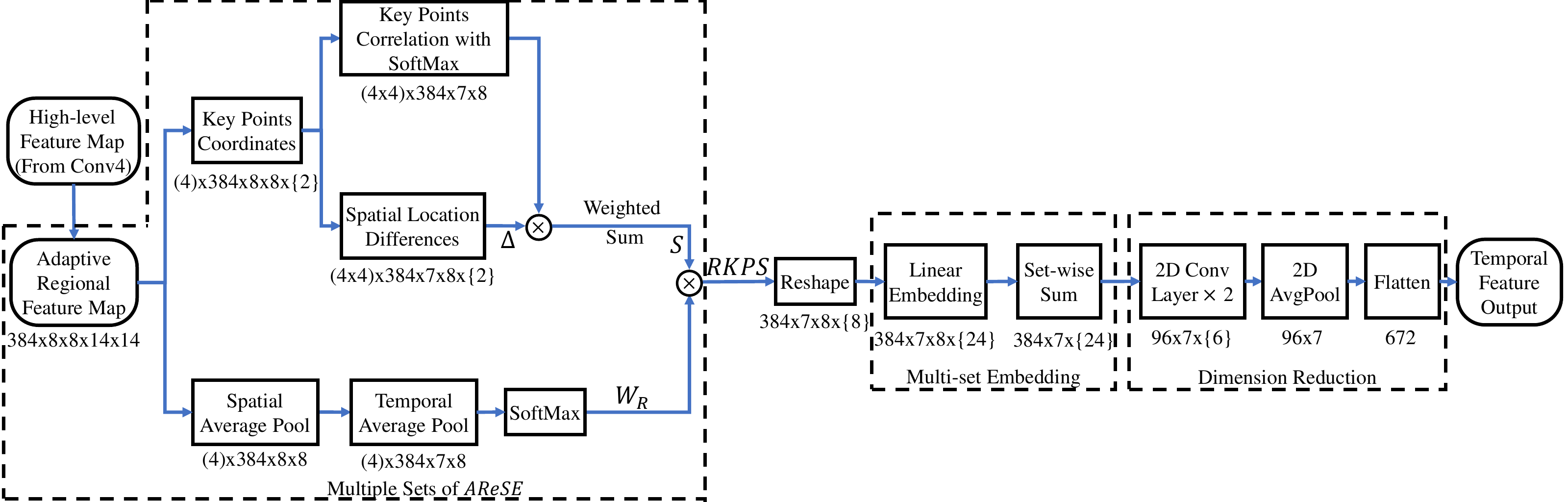}}\\
    \caption{Detailed implementation of (a) our action recognition framework MF-KPSEM, utilizing MFNet as the backbone for spatial feature extraction, with our proposed $KPSEM$ for temporal feature extraction; and (b) more detailed implementation of $KPSEM$ with 8 sets of $AReSE$s. For (a), the figures above or below the blocks are the output size of the respective blocks expressed in $C\times T\times H\times W$. Here C is the number of channels, T, H and W are the length of time, height and width of the features. For (b), the figures below the blocks are the output size of the respective blocks. The number within the parentheses corresponds to the dimension reflecting the split of frames $K$, while the number within the braces corresponds to the key point shifts or its embedding. The other figures are expressed in the order of $C\times T\times G\times H\times W$. Here $C$ is the number of channels, $T$, $G$, $H$ and $W$ are the length of time, number of sets for $AReSE$s (\textit{cf.} Section \ref{section:AReSE}), height and width of the features. The black path is the path to extract spatial features with MFNet. Whereas the blue path is the path to extract temporal features with $KPSEM$. Figures are best viewed zoomed in.}
    \label{detail:all}
\end{figure*}

As mentioned in Section \ref{section:method}, our proposed $KPSEM$ utilizes MFNet as the backbone CNN network. Here we present a more detailed implementation of the overall framework of MF-KPSEM, shown in Figure~\ref{detail:framework}, and the detailed implementation of our proposed $KPSEM$ as shown in Figure~\ref{detail:KPSEM}.

We follow the implementation in \cite{chen2018multifiber} where the input is a frame sequence of 16 frames, with each frame of size $224\times224$. As mentioned in Section \ref{section:overall} of our paper, the input of $KPSEM$ is the output of Conv4 layer of MFNet~\cite{chen2018multifiber}, which is of size $384\times8\times14\times14$. To obtain the temporal feature output, $KPSEM$ first computes the 8 $RKPS$s obtained through 8 sets of $AReSE$s. Each set of $AReSE$s performs different region separations, splitting the high-level feature maps from the output of Conv4 layer adaptively. All the $RKPS$s go through a reshape process before going through the multi-set embedding operation. Each $RKPS$ is linearly projected separately to a dimension size of 24 as mentioned in Section \ref{experiment:settings}. To obtain the overall feature output, a series of dimension reduction operation is utilized after the set-wise summation of the linear embeddings. Specifically, two 2D convolution layers with a kernel size of $(1,3)$ and a stride $(1,2)$ are first utilized to reduce the embedding from the size of $24$-d to $6$-d. Subsequently, the result passes through a 2D average pooling layer and then flattened to produce the final temporal feature output. It is worth noticing that in addition to reducing embedding size, we also reduce the channels from the size of $384$-d to $96$-d to discard the non-salient embeddings. The size of the temporal feature output is of $672$-d.

\subsection{Results and Comparison}
Table \ref{table:compare-sota} shows the comparison of top-1 accuracy on Mini-Kinetics, UCF101, Something-Something v1 and HMDB51 datasets with other state-of-the-art methods including:
\begin{enumerate}
    \item \textit{Two-stream CNN methods: }MARS \cite{crasto2019mars}, Residual Frame with two-stream input (ResFrame TS) \cite{tao2020rethinking} and I3D with two-stream input (I3D TS) \cite{carreira2017quo}.
    \item \textit{2D CNN \& 3D CNN methods: }C3D \cite{tran2015learning}, I3D with RGB input \cite{carreira2017quo}, (2+C1)D \cite{cheng2019sparse}, S3D \cite{xie2018rethinking}, MFNet \cite{chen2018multifiber}, ECO \cite{ECO_eccv18}, ${\rm ECO}_{Lite}$ \cite{ECO_eccv18}, TSM \cite{Lin_2019_ICCV} and TSN \cite{wang2016temporal}. 
    \item \textit{CNN with learnable feature correlations: }TBN \cite{li2019temporal}, Res50-NL \cite{wang2018non}, Res50-CGD \cite{he2019compact}, Res50-CGNL \cite{yue2018compact}, I3D-NL \cite{wang2018non} and I3D-NL-GCN \cite{wang2018videos}.
\end{enumerate}
\label{experiment:results-comparion}

\begin{table*}[!ht]
  \centering
  \small
  \setlength{\tabcolsep}{4pt}
  \resizebox{.9\textwidth}{!}{
  \smallskip\begin{tabular}{c|c|c|c|c|c|c|c}
    \hline
    \hline
    & Method & Mini-Kinetics & UCF101 & STH-STH v1 & HMDB51 & \# Params & FLOPs\\
    \hline
    \multirow{3}{*}{\parbox{3cm}{\centering Two-stream CNNs}}
    & MARS \cite{crasto2019mars}        & 73.5\% & \textbf{98.1}\% & \textbf{53.0\%} & \textbf{80.9}\% &- & -\\
    & ResFrame TS \cite{tao2020rethinking}   & 73.9\% & 90.6\% & - & 55.4\% & - & -\\
    & I3D (TS) \cite{carreira2017quo}      & 78.7\% & 97.9\% & - & 80.2\% & 25.0M & $>\!107.9$G\\
    \hline
    \multirow{9}{*}{2D CNNs \& 3D CNNs}
    & C3D \cite{tran2015learning}        & 66.2\% & 85.2\% & - & - & 33.3M & -\\
    & I3D (RGB) \cite{carreira2017quo}     & 74.1\% & 95.4\% & 45.8\% & 74.5\% & 12.06M & 107.9G\\
    & (2+C1)D \cite{cheng2019sparse}      & 75.74\% & 96.9\% & - & 75.2\% & \textbf{7.3M} & 31.9G\\
    & S3D \cite{xie2018rethinking}       & 78.0\% & 96.8\% & 48.2\% & 75.9\% & 8.77M & 43.47G\\
    & MFNet \cite{chen2018multifiber}      & 78.35\% & 96.0\% & 43.0\% & 74.6\% & 8.0M & \textbf{11.1G}\\
    & ECO \cite{ECO_eccv18}           & - & 94.8\% & 41.4\% & 72.4\% & 47.5M & 64G\\
    & ${\rm ECO}_{Lite}$ \cite{ECO_eccv18}   & - & - & 46.4\% & - & 150M & 267G\\
    & TSM \cite{Lin_2019_ICCV}      & - & 95.9\% & 49.7\% & 73.5\% & 48.6M & 98G\\
    & TSN \cite{wang2016temporal}      & - & 94.2\% & 19.5\% & 69.4\% & 10.7M & 16G\\
    \hline
    \multirow{6}{*}{\parbox{3cm}{\centering CNN with learnable feature correlations}}
    & TBN \cite{li2019temporal}         & 69.5\% & 93.6\% & - & 69.4 & 11.4M & -\\
    & Res50-NL \cite{wang2018non}        & 77.53\% & 82.88\% & - & - & 27.66M & 19.67G\\
    & Res50-CGD \cite{he2019compact}      & 77.56\% & 84.06\% & - & - & 25.58M & 17.88G\\
    & Res50-CGNL \cite{yue2018compact}     & 77.76\% & 83.38\% & - & - & 27.2M & 19.16G\\
    & I3D-NL \cite{wang2018non}            & - & - & 44.4\% & - & 27.2M & 19.16G\\
    & I3D-NL-GCN \cite{wang2018videos}  & - & - & 46.1\% & - & 27.2M & 19.16G\\
    \hline
    Ours & \textbf{MF-KPSEM}           & \textbf{82.05\%} & 97.4\% & 48.1\% & 77.7\% & 8.11M & 11.21G\\
    \hline
    \hline
  \end{tabular}
  }
  \caption{Comparison of top-1 accuracy, number of parameters and computational cost in FLOPs with state-of-the-art methods on Mini-Kinetics, UCF101, Something-Something v1 and HMDB51 datasets. MF-KPSEM instantiates MFNet as the backbone CNN.}
  \smallskip
  \label{table:compare-sota}
\end{table*}

Our state-of-the-art performance is achieved by instantiating MFNet, denoted as MF-KPSEM. For the experiments as presented in Table \ref{table:compare-sota}, the batch size is set to 64 for Mini-Kinetics as well as Something-Something v1 datasets, 80 for UCF101 dataset and 128 for HMDB51 dataset, respectively. The experiments are conducted using two NVIDIA Quadro RTX8000 GPUs. 

\begin{figure*}[!t]
  \centering{
  \includegraphics[width=.8\textwidth]{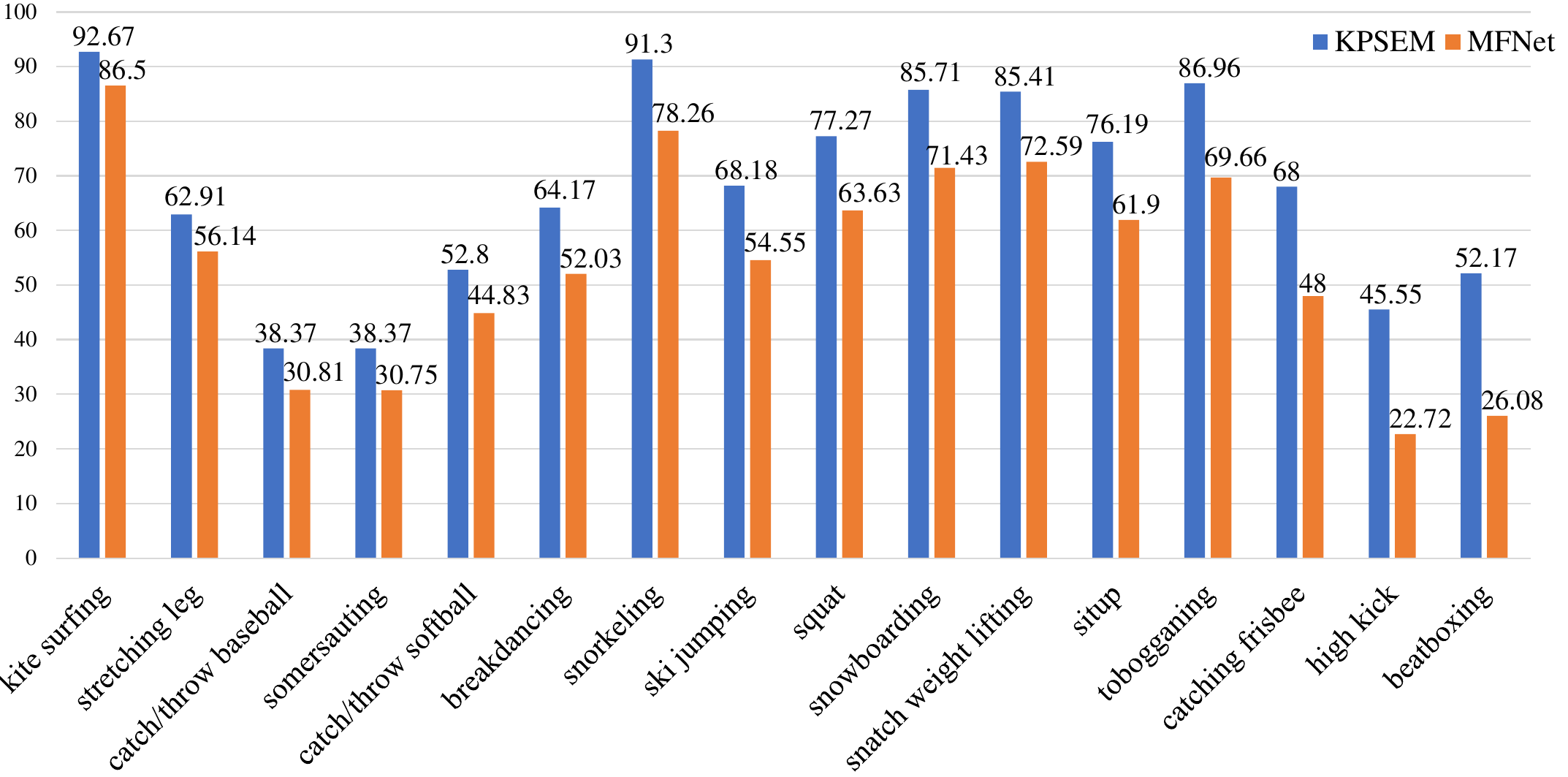}}
  \hfill
  \caption{Detailed comparison of accuracy per class on Mini-Kinetics. Here we present the accuracies of 16 classes where MF-KPSEM outperforms by a margin of at least $6\%$. }
  \label{detail:compare-classes}
\end{figure*}
\begin{figure*}[!t]
  \centering{
  \includegraphics[width=.8\textwidth]{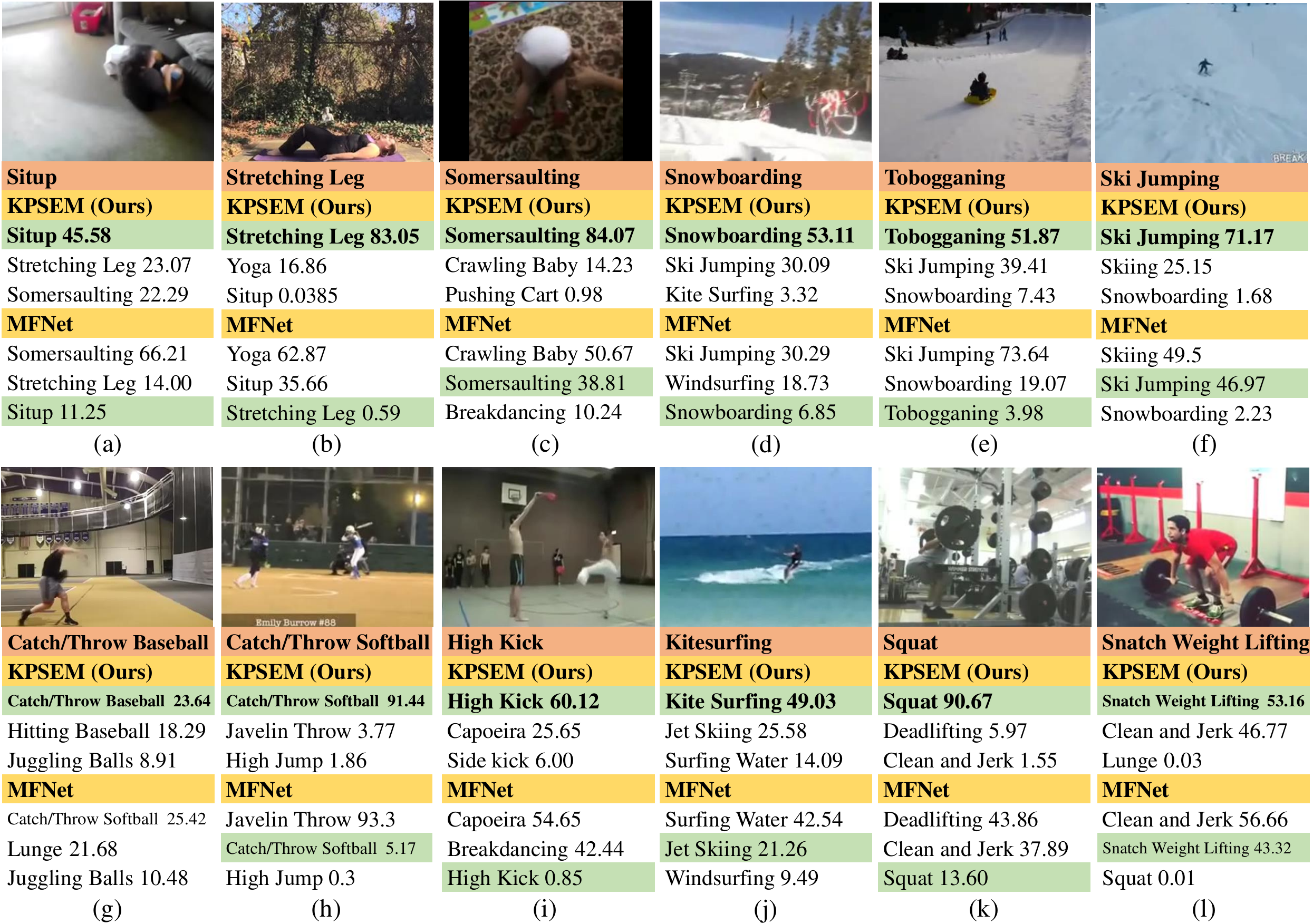}}
  \hfill
  \caption{Twelve examples taken from the 16 classes presented in Figure~\ref{detail:compare-classes}. The numbers on the right of each class show the probability of the class from the classifier in percentages. We show three classes with the highest probability. The class with the highest probability is the result of the top-1 classification.}
  \label{detail:illustrate-examples}
\end{figure*}

The performance results in Table \ref{table:compare-sota} show that our network achieves the state-of-the-art result on the Mini-Kinetics with only a minor increase in the number of parameters and required computational cost. Our proposed MF-KPSEM achieves a relative $4.73\%$ increase in recognition accuracy over our baseline method, at a cost of a mere $1.38\%$ increase in parameters and $1\%$ extra FLOPs. Our method surpasses the previous state-of-the-art method which utilizes learnable feature correlations through the CGNL module by $4.29\%$, with a much lighter network and requires lower computational cost. It is noted that the number of parameters in MF-KPSEM exceeds that of (2+C1)D, which is built on top of DenseNet \cite{huang2017densely} that is characterized by its small parameter size. However, our method exceeds theirs by $6.31\%$ in accuracy with $64.86\%$ reduced FLOPs. 

For the other three datasets, our MF-KPSEM also performs competitively, gaining a $1.4\%$ increase for UCF101, $5.1\%$ increase for Something-Something v1 and $3.1\%$ for HMDB51 compared to our baseline network. For UCF101 and HMDB51 datasets, our proposed MF-KPSEM performs slightly poorer than the two-stream CNN MARS \cite{crasto2019mars}, trailing by $0.7\%$ for UCF101 and $3.2\%$ for HMDB51. Yet our approach does not utilize optical flow as our input during both training and inference, which means a significant reduction in memory and computational cost. For Something-Something v1, while our MF-KPSEM performs slightly inferior compared to S3D \cite{xie2018rethinking} and TSM \cite{li2019temporal}, our MF-KPSEM is a much lighter network with smaller parameter size and requires less computational cost. Specifically, the parameters of our MF-KPSEM are $7.5\%$ and $83.3\%$ less than those of S3D and TSM, respectively. For computational cost, MF-KPSEM requires $74.2\%$ fewer FLOPS compared to S3D and $88.6\%$ fewer FLOPS compared to TSM. Despite the obvious gap of parameters and FLOPS, our MF-KPSEM still performs competitively for Something-Something v1, with a minor $0.1\%$ and $1.8\%$ gap in accuracy compared to S3D and TSM, respectively.

We further investigate the improvement over different actions and present a more detailed comparison of performance between our proposed MF-KPSEM network and the baseline MFNet network. Figure~\ref{detail:compare-classes} shows the accuracy of 16 classes from the Mini-Kinetics dataset, where our network outperforms the original network by a noticeable margin of over $6\%$. Many of the examples in these classes are characterized by the fact that the frames in each of these examples would appear similar to other action classes. Therefore, the temporal features showing how the action evolves is the key to correctly classifying these examples. 

Figure~\ref{detail:illustrate-examples} present 12 examples from the 16 classes mentioned in Figure~\ref{detail:compare-classes}, all of which are better classified through the proposed MF-KPSEM. It could be observed that the spatial features of the given examples, or more intuitively the appearance of the frames in the given examples, could not provide effective representation for accurate action recognition. For example, for Video (a) in Figure~\ref{detail:illustrate-examples}, the actor is seen rolling up herself. Such a scenario could be present in the action class ``Somersaulting", in which actors would roll themselves up to turn upside down. It could also be presented in the action class ``Situp", where the rolling up is followed by the actor rolling backwards to roll up again. The class of this action could only be determined through the temporal features, which are the change of the actor's position. In this video, the actor is rolling backwards after this scenario. The actual change of the actor's position suggests that the action should be classified as ``Situp". If only spatial features are utilized as in the case of MFNet, the video would be instead classified as ``Somersaulting" due to the multiple frames showing the actor rolled up. This clearly shows the importance of temporal features in accurate action recognition and the effectiveness of our $KPSEM$ in extracting effective temporal features.

\begin{table}[!ht]
  \centering
  \small
  \subfloat[Separation of Feature Maps\label{table:ablation-1}]{
  \resizebox{.3\textwidth}{!}{
  \smallskip\begin{tabular}{c|c}
    \hline
    \hline
    Model & Accuracy\\
    \hline
    MFNet & 70.8\%\\
    \hline
    MF-KPSEM & 74.0\%\\
    Single Key Point & 72.1\%\\
    \hline
    \hline
  \end{tabular}
  }
  \hspace{1em}
  }
  \subfloat[Adaptive Regions\label{table:ablation-2}]{
  \resizebox{.3\textwidth}{!}{
  \smallskip\begin{tabular}{c|c}
    \hline
    \hline
    Model & Accuracy\\
    \hline
    MFNet & 70.8\%\\
    \hline
    MF-KPSEM & 74.0\%\\
    Fixed Regions & 72.5\%\\
    \hline
    \hline
  \end{tabular}
  }
  } \\
  \subfloat[Multiple Sets of $AReSE$\label{table:ablation-3}]{
  \resizebox{.3\textwidth}{!}{
  \smallskip\begin{tabular}{c|c}
    \hline
    \hline
    Model & Accuracy\\
    \hline
    MFNet & 70.8\%\\
    \hline
    MF-KPSEM & 74.0\%\\
    One Set of $AReSE$ & 72.4 \\
    \hline
    \hline
  \end{tabular}
  }
  \hspace{1em}
  }
  \subfloat[$KPSEM$ with R3D\label{table:ablation-4}]{
  \resizebox{.3\textwidth}{!}{
  \smallskip\begin{tabular}{c|c}
  \hline
    \hline
    Model & Accuracy\\
    \hline
    R3D \cite{hara2018can}& 62.0\%\\
    R3D-KPSEM & 65.8\%\\
    \hline
    \hline
  \end{tabular}
  }
  }\\
  \caption{Ablations of $KPSEM$ utilizing MFNet on HMDB51 split-1. The ablation is performed with a small batch size of 16. The networks with variants of $KPSEM$ that utilize (a) only 1 single key point per frame, (b) only fixed regions per frame and (c) with only 1 set of $AReSE$ is compared with the proposed $KPSEM$, and (d) different backbone networks.}
  \label{table:ablations} 
\end{table}

\subsection{Ablation Study and Visualization}
\label{experiment:ablation-visualization}

In this section, we justify our proposed design of $KPSEM$ through ablation study and visualization of results. Specifically, we examine the performance of our $KPSEM$ in four scenarios and justify the need for high-level feature maps as input for $KPSEM$, multiple key points per frame, adaptive regions in each frame and multiple sets of $AReSE$. Additionally, we combine our $KPSEM$ with another baseline network R3D to justify the robustness of $KPSEM$. We further examine the effectiveness of $KPSEM$ by visualizing heatmaps and the extracted key points with their corresponding shift. The split-1 of HMDB51 dataset is adopted for all ablation studies, trained with a batch size of 16 on a single NVIDIA TITAN Xp GPU.

\paragraph{Position of $KPSEM$}
Our proposed $KPSEM$ module utilizes high-level feature maps as the input for extracting key point shifts. Figure~\ref{exp:settings-1} compares the result when $KPSEM$ is added to different stages of MFNet. $KPSEM$ is added right after the respective layers. Though improvements have been made for all networks utilizing $KPSEM$ regardless of the position, the improvement achieved when $KPSEM$ is added after Conv2 is $1.5\%$ smaller than that when $KPSEM$ is added after Conv4. One possible explanation is that the representation level of the feature maps after Conv2 layer is lower than that of the feature maps after Conv4 layer. This indicates that points with higher feature values from Conv2 layer may be more relevant to pixels with higher values rather than key points. Pixels with higher values may not be key points as they may correspond to the white background pixels. This indicates that the key point shifts extracted from Conv2 feature maps may be semantically less representative than those extracted from Conv4 feature maps, thus resulting in inferior performances. We also observe a sharp drop in performance when $KPSEM$ is positioned after Conv5 layer. We believe that this is because of the small spatial size of Conv5 feature maps ($7\times7$). The Conv5 feature maps are therefore insufficient to provide spatial information of the key points, decreasing the accuracy of the extracted key point shifts. This ends up in the inferior performance when $KPSEM$ is positioned after Conv5 layer.

\paragraph{Separation of Feature Maps}
As mentioned in Section \ref{section:KPSEM}, frames are split such that localized key points distributed across each frame are preserved. We justify the need for multiple key points per frame by comparing with the variant of $KPSEM$ where only a single global key point utilized for each frame at each channel. As indicated in Table \ref{table:ablation-1}, the use of multiple key points (in this case 4 key points) per frame boosts the performance by $1.9\%$. This demonstrates the effectiveness of extracting multiple regional key points across each frame. The results are consistent with that shown in Figure~\ref{exp:settings-2}, where all results are higher than the result using only a single key point.

However, more splits per frame do not guarantee higher accuracy. The result in Figure~\ref{exp:settings-2} shows that splitting each frame with $K=9$ or $K=16$ regions results in a slight decrease in performance compared to that of splitting $K=4$ regions per frame. The slightly worse performance for higher $K$ settings could be explained by higher region number results in smaller regions, which may result in regions with only the background being split out. The key points extracted from regions corresponding to the background and their related key point shifts would be redundant. The resulting temporal features would therefore be less effective and result in inferior classification accuracies.

\paragraph{Adaptive Separation of Feature Maps}
Also mentioned in Section \ref{section:KPSEM}, fixed splits for each frame may result in different key points located within the same fixed region. To mitigate this drawback of fixed splits, we split each frame adaptively to obtain an optimal region separation solution for key point extraction. We justify the need for adaptive splits by comparing with the variant of $KPSEM$ utilizing fixed splits for each frame. For this variant of $KPSEM$, the split of each set of $AReSE$ is randomly initialized, while other network settings including the number of sets and the number of splits remain the same as the default settings of our proposed $KPSEM$. After initialization, the splits of each frame are fixed during the training process. Results in Table \ref{table:ablation-2} demonstrate that a $1.5\%$ increase in accuracy is achieved through splitting frames adaptively. This validates that adaptive splitting could result in better temporal features.

\paragraph{Multiple Sets of $AReSE$}
To extract more key points while avoiding over splitting each frame, we proposed to use multiple sets of $AReSE$ in Section \ref{section:KPSEM} and Section \ref{section:AReSE}. We examine the effect of using multiple sets of $AReSE$ with its results as presented in Table \ref{table:ablation-3}. The use of 8 sets of $AReSE$ helps improve the network by $1.6\%$ over its variant which uses only a single set of $AReSE$. This matches the results shown in Figure~\ref{exp:settings-3}, where the accuracy rises as the number of sets of $AReSE$ increases in general. All results using multiple $AReSE$ outperform that of using a single set of $AReSE$. The increase justifies the effectiveness of increasing extracted key points without over splitting each frame.

As mentioned in \ref{section:KPSEM}, the number of $RKPS$ increases with the increasing number of sets of $AReSE$. For $G\leqslant8$, a significant increase in accuracy with an increase of $G$ can be observed. This suggests that the increase in set number $G$ results in different $RKPS$, or key point shifts, extracted under different region separations. The different key point shifts extracted by multiple sets of $AReSE$ constitutes more robust temporal features. It is worth noticing that the accuracy saturates for $G>8$. This suggests that a further increase in $G$ may not result in different region separations. The increased $RKPS$s may be a repetition of previous $RKPS$s, which therefore may not increase the robustness of temporal features and the classification accuracy.

\paragraph{$KPSEM$ with Other Backbones}
Our proposed $KPSEM$ can be applied with any other CNN backbones and improve their performance. To demonstrate the robustness of our $KPSEM$, we have conducted another experiment on $KPSEM$ with a different backbone, namely the 3D variant of ResNet50 \cite{hara2018can} referred to as R3D \cite{wang2018non}. R3D is constructed by inflating 2D convolution kernels directly to 3D convolution kernels, implemented as $1\times k \times k$ kernels. Therefore, the R3D simply aggregates the input frames and can be directly initialized from weights pretrained on ImageNet. Our $KPSEM$ is inserted after the res3 layer of R3D which results in a $3.8\%$ improvement for top-1 accuracy compared to the baseline. The improvement which $KPSEM$ achieves with R3D as well as MFNet \cite{chen2018multifiber} justifies the robustness of our proposed $KPSEM$ module.

\begin{figure*}[!ht]
  \centering
  \subfloat[\label{visual:wave}]{
  \includegraphics[width=.4\textwidth]{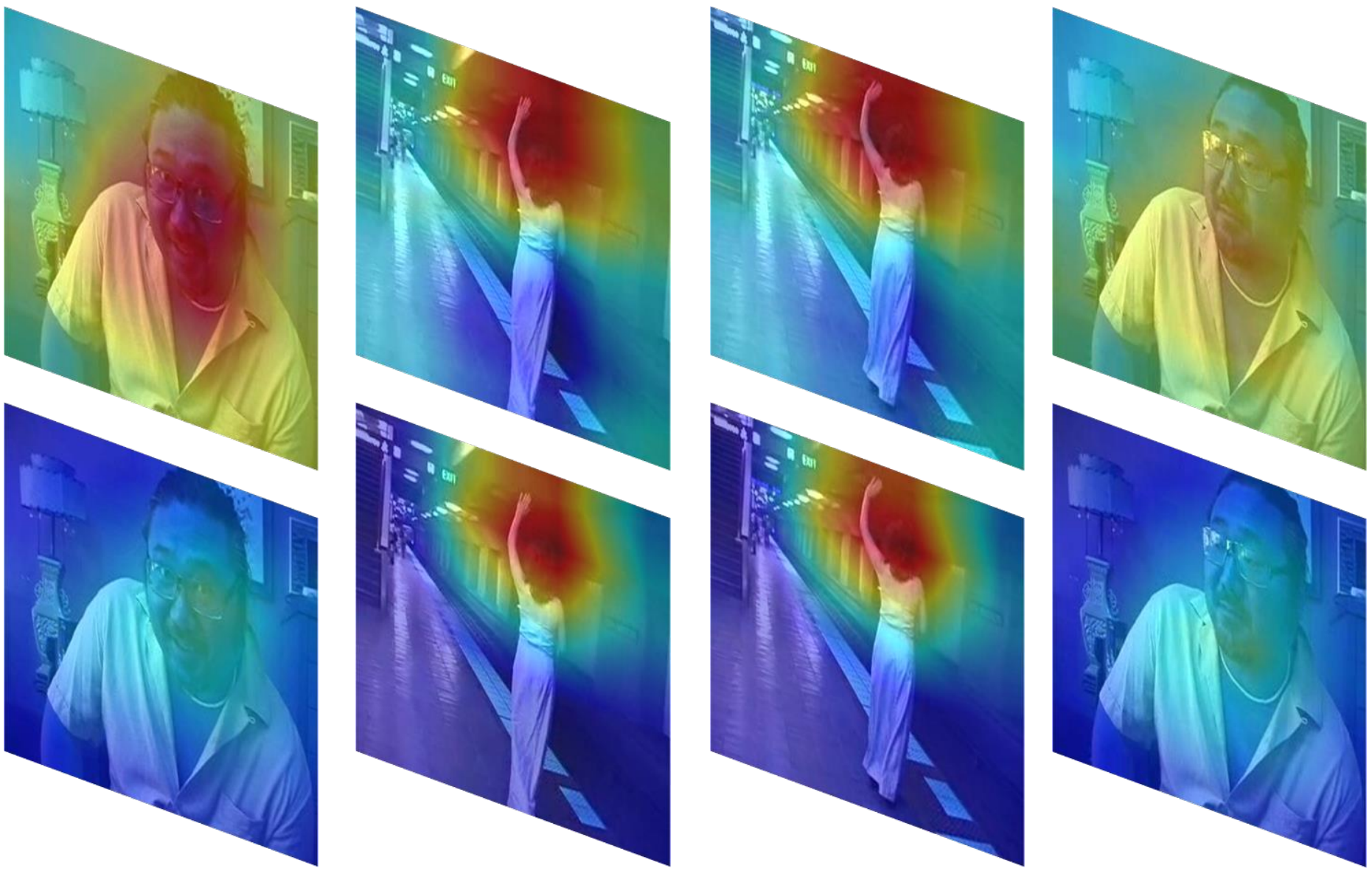}}
  \hspace{.5em}
  \subfloat[\label{visual:turn}]{
  \includegraphics[width=.4\textwidth]{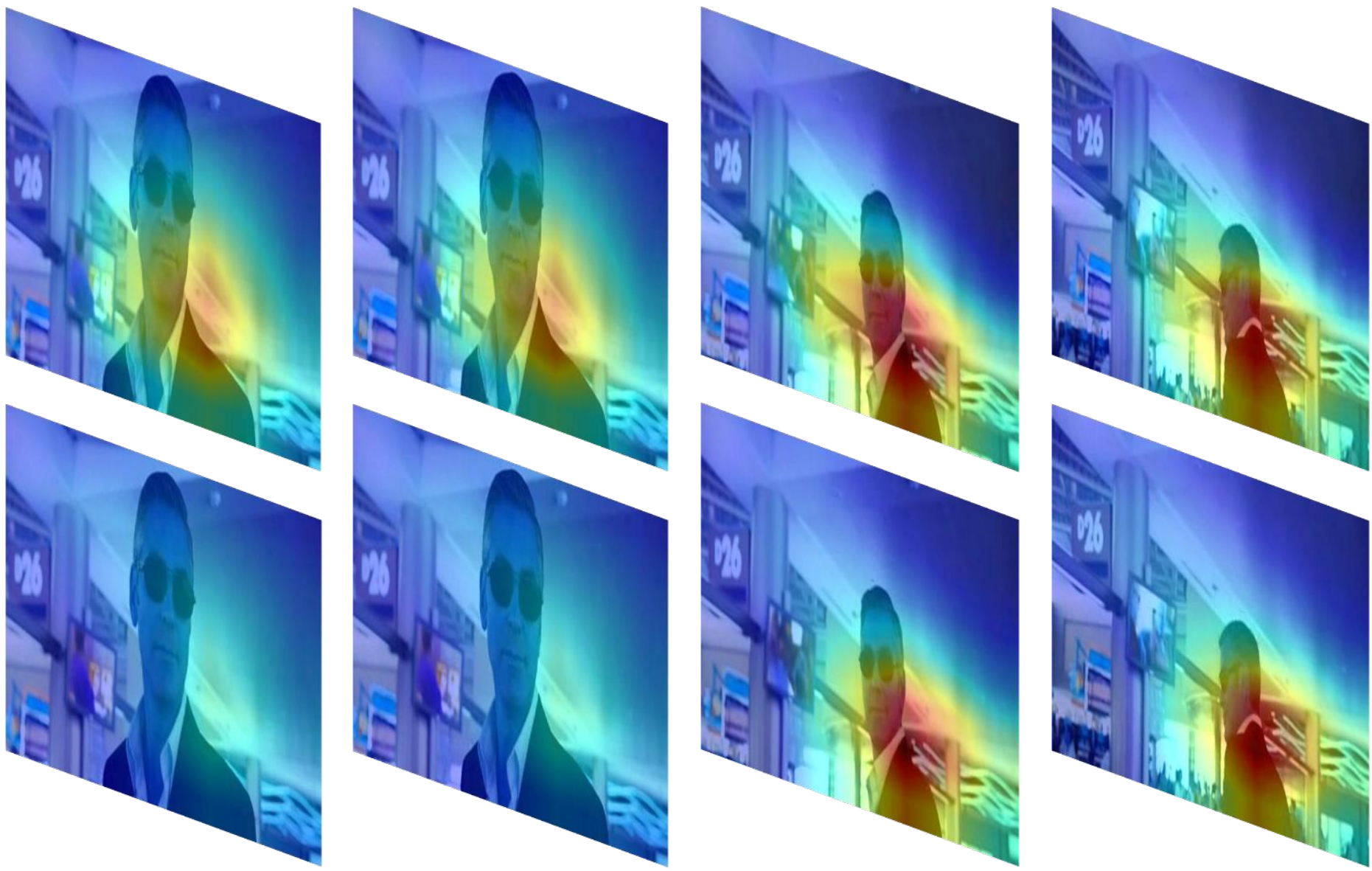}}
  \\[1ex]
  \subfloat[\label{visual:swingbaseball}]{
  \includegraphics[width=.4\textwidth]{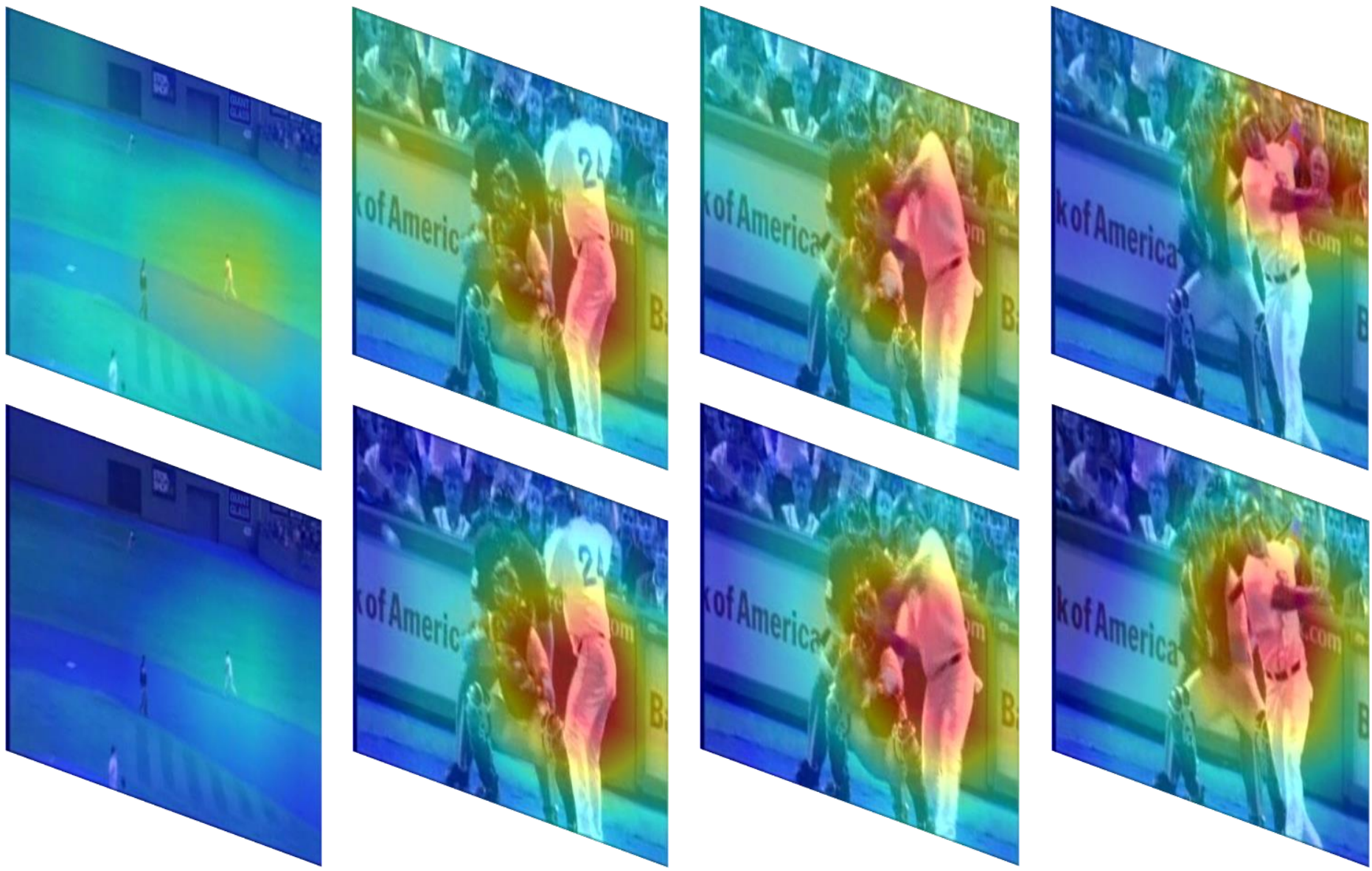}}
  \hspace{.5em}
  \subfloat[\label{visual:sword}]{
  \includegraphics[width=.4\textwidth]{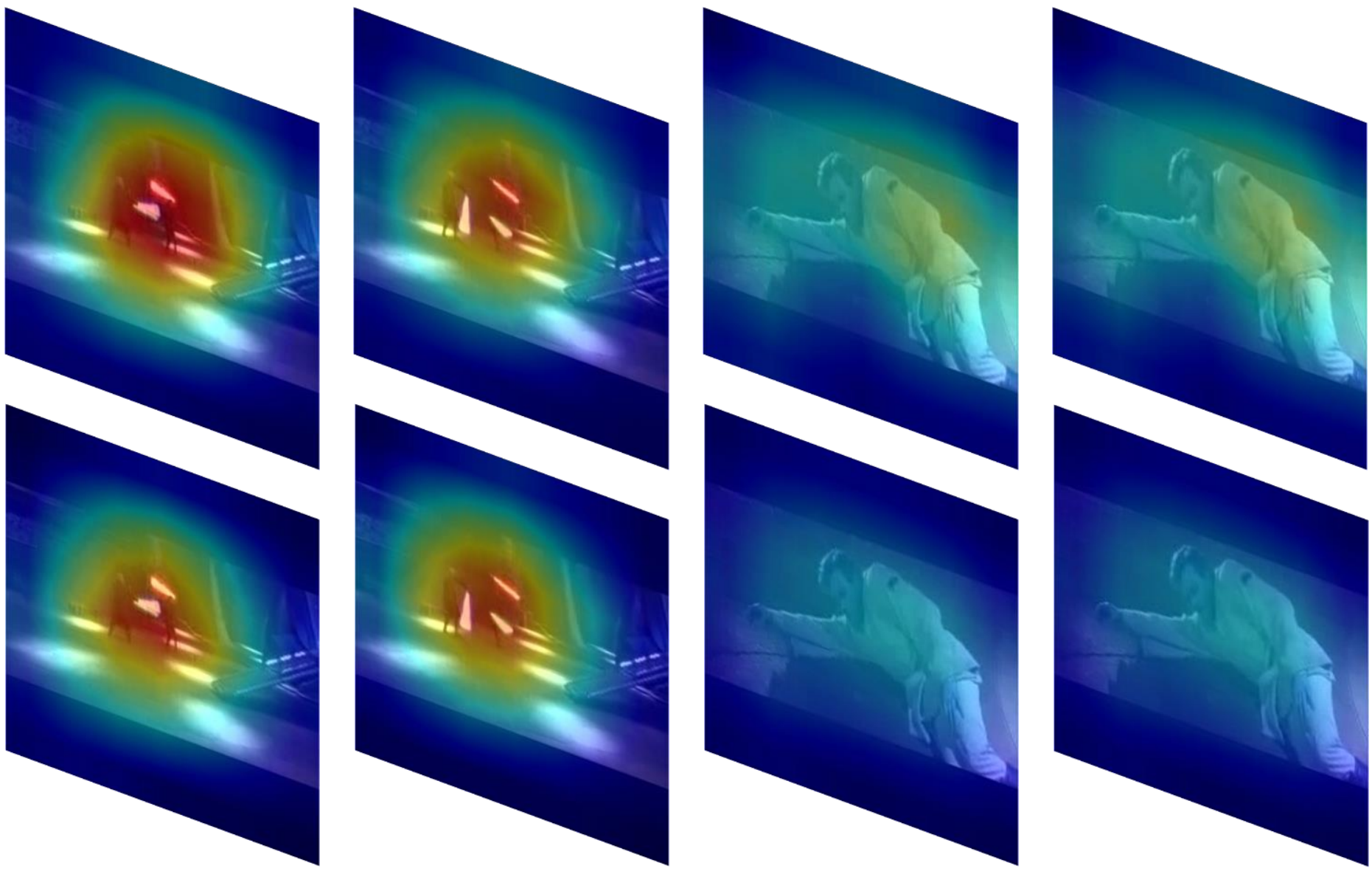}}\\
  \caption{Heatmaps of four samples of HMDB51. For each sub-figure, the upper is the result of baseline MFNet and the lower is the result of our proposed MF-KPSEM. Figure~\ref{visual:wave}, Figure~\ref{visual:turn}, Figure~\ref{visual:swingbaseball} and Figure~\ref{visual:sword} belong to action classes ``Wave", ``Turn", ``Swing baseball" and ``Sword", respectively. Compared to the baseline, our proposed MF-KPSEM accurately concentrates on the key regions which are relevant to the ground-truth actions, while the baseline model might focus on irrelevant regions instead.}
  \label{visual:CAM}
\end{figure*}

\begin{figure*}[!ht]
  \centering
  \subfloat[Visualizing Action ``Climbing"\label{detail:visualize-1}]{
    \includegraphics[width=.42\textwidth]{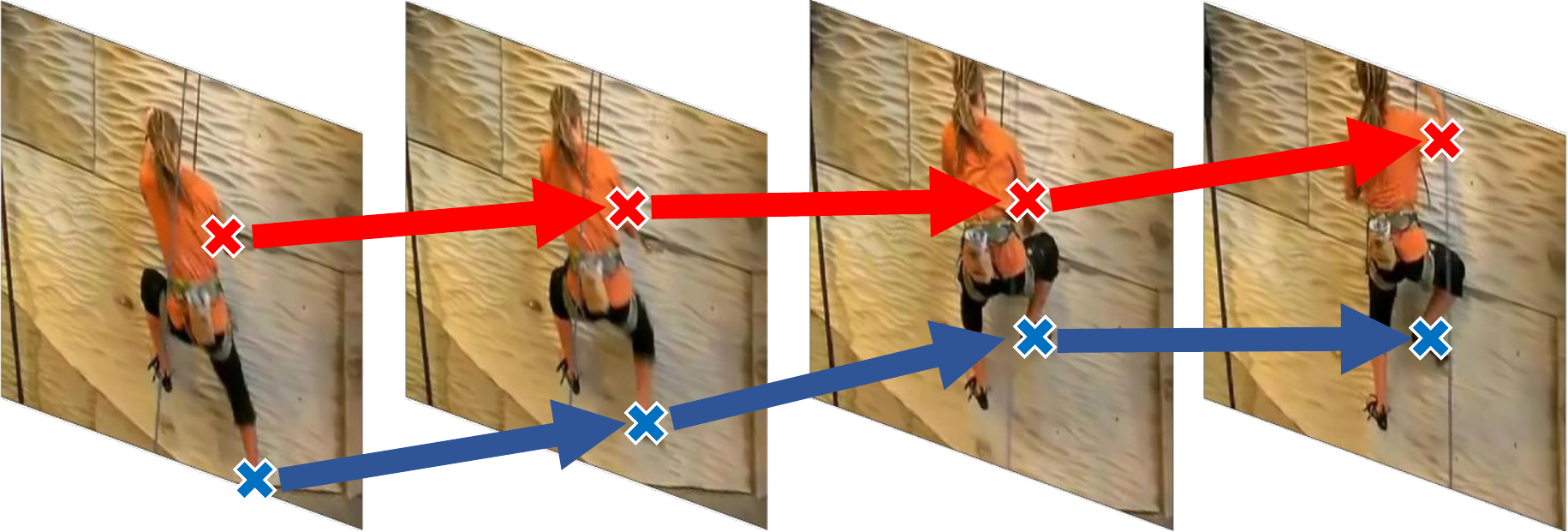}}
    \hfill
    \subfloat[Visualizing Action ``Archery"\label{detail:visualize-2}]{
    \includegraphics[width=.42\textwidth]{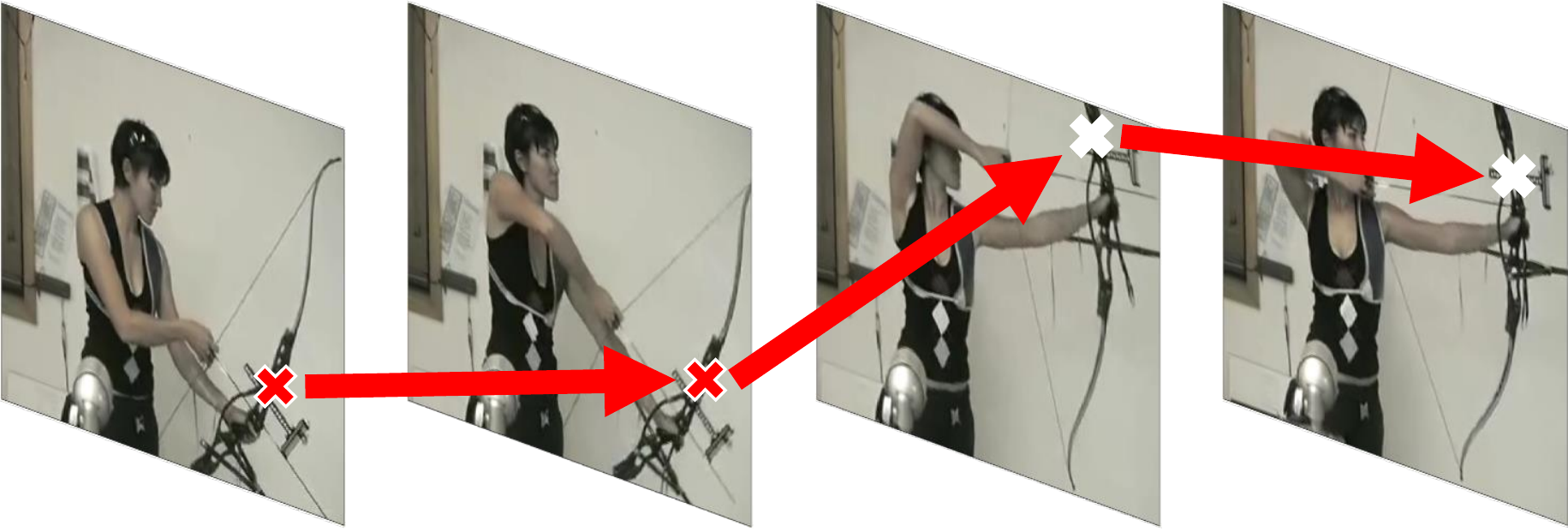}}
    \hfill
    \subfloat[Visualizing Action ``Sit-up"\label{detail:visualize-3}]{
    \includegraphics[width=.42\textwidth]{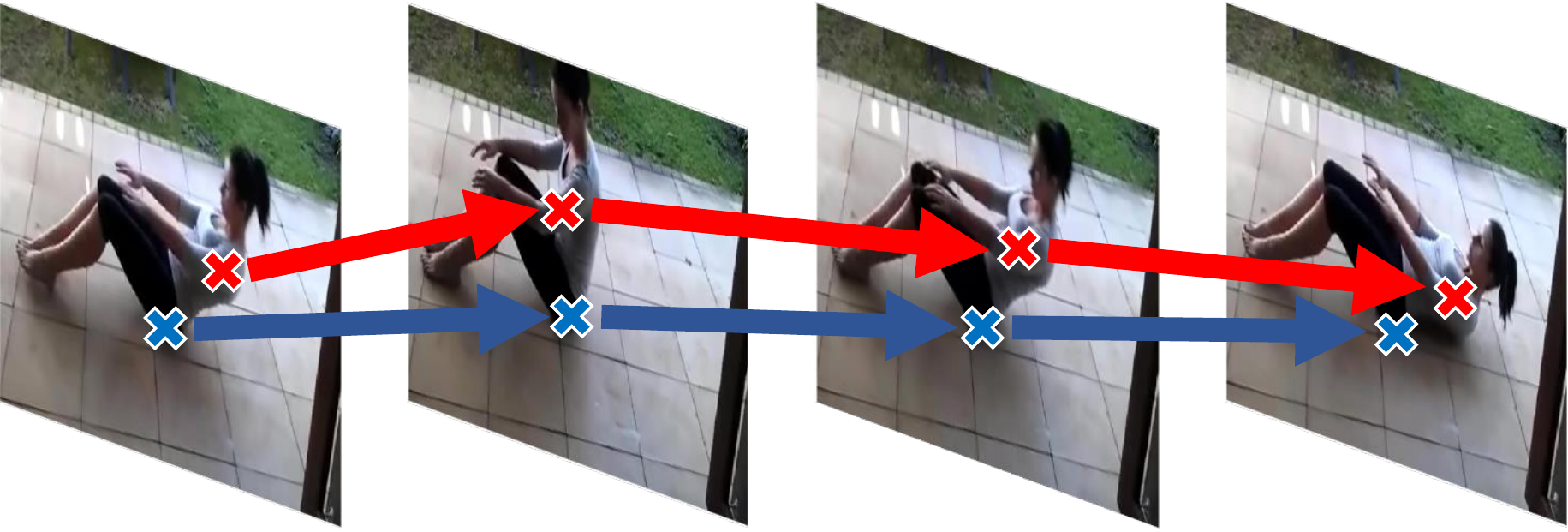}}
    \hfill
    \subfloat[Visualizing Action ``High Kick"\label{detail:visualize-4}]{
    \includegraphics[width=.42\textwidth]{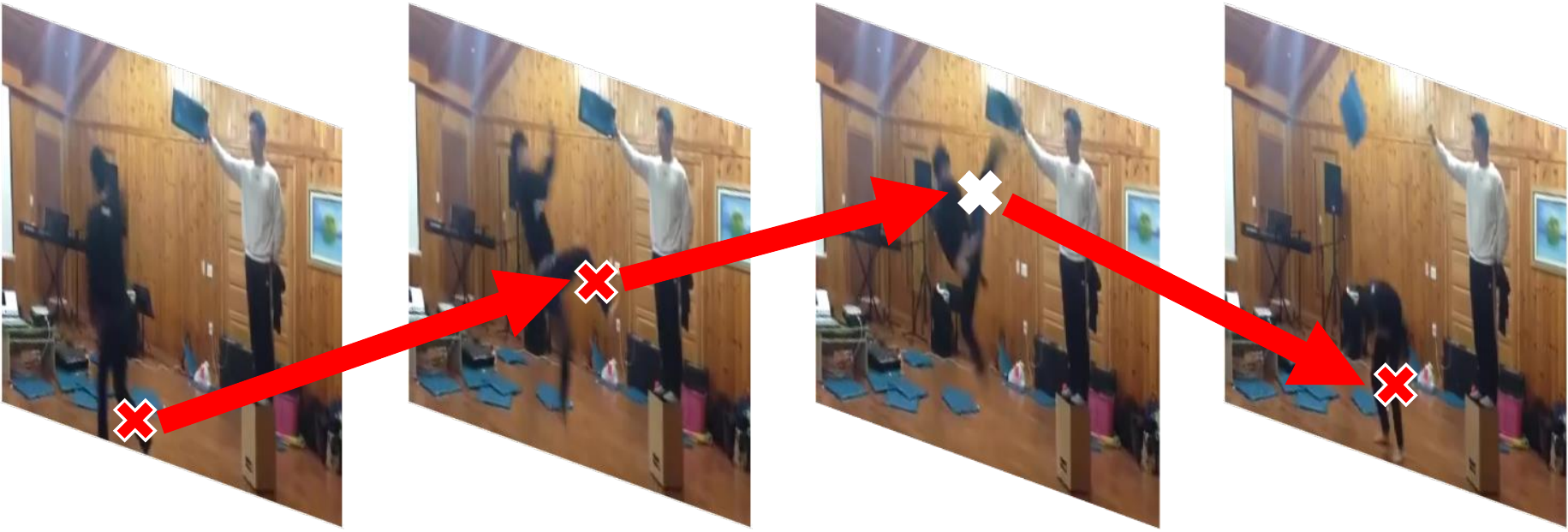}}
    \hfill
    \subfloat[Visualizing Action ``Kick Ball"\label{detail:visualize-5}]{
    \includegraphics[width=.42\textwidth]{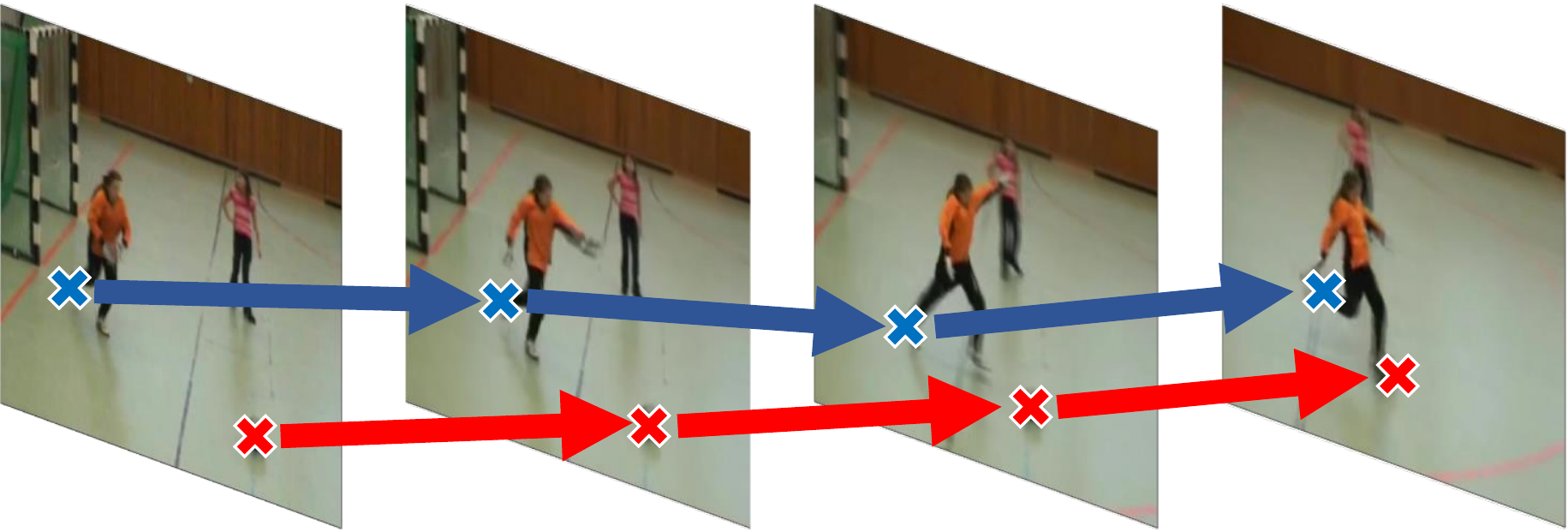}}
    \hfill
    \subfloat[Visualizing Action ``Tobogganing"\label{detail:visualize-6}]{
    \includegraphics[width=.42\textwidth]{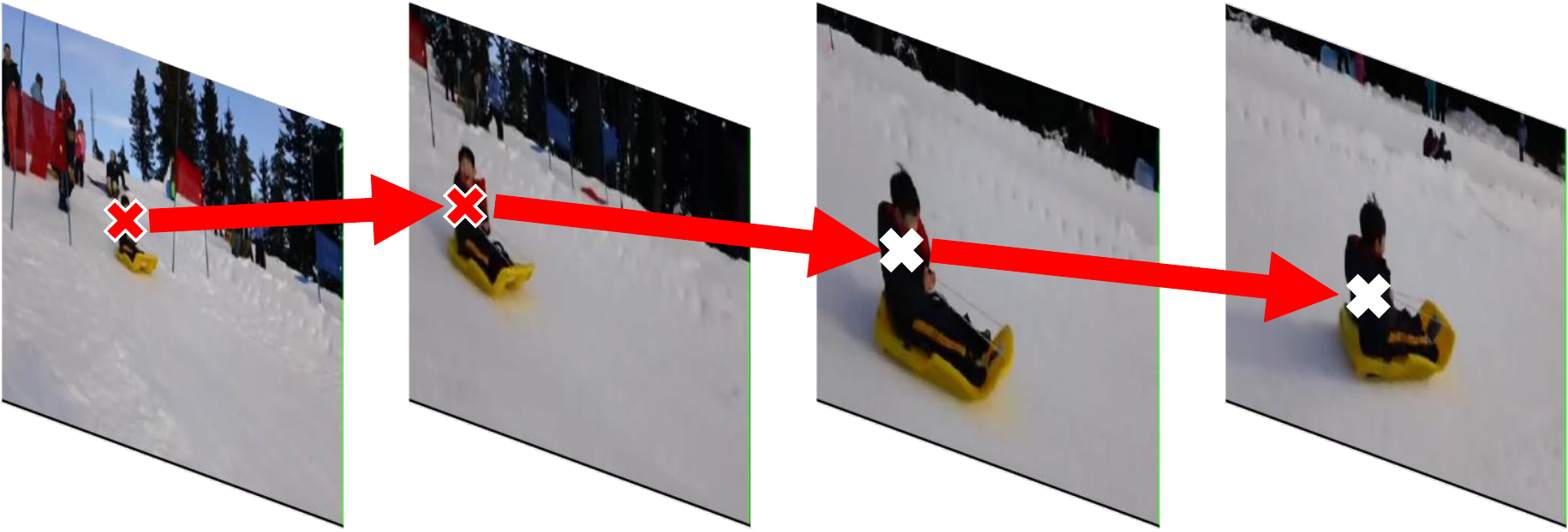}}\\
  \caption{Extracted key points and their corresponding shift from MF-KPSEM for six videos. Crosses coloured differently indicate that they are located in different regions. Arrows in different colours indicate the respective key point shifts. Figure~\ref{detail:visualize-1}, Figure~\ref{detail:visualize-2} and Figure~\ref{detail:visualize-5} are videos from HMDB511 dataset and the others are from Mini-Kinetics. Figure best viewed in colour and zoomed in.}
  \label{detail:visualize-sup}
\end{figure*}

\paragraph{Heatmaps Visualization}
To investigate where the MF-KPSEM and the baseline MFNet focus on, we visualize the heatmaps that indicate the focus of either network on sampled frames. Specifically, the heatmaps are computed based on Gradient-weighted Class Activation Mapping (Grad-CAM) \cite{Selvaraju_2017_GradCAM}. The Grad-CAM results are extracted from the last convolution layer of both MF-KPSEM and the baseline MFNet. Figure~\ref{visual:CAM} illustrates Grad-CAM results of four examples from the test set of split 1 of HMDB51, each of which includes four frames sampled identically from the input 16 frames. For each sub-figure, the upper sequence of frames is extracted from the baseline MFNet while the lower sequence is from MF-KPSEM. It can be observed that our proposed MF-KPSEM focuses on the key regions relevant to the action more accurately compared to the baseline model. Whereas the baseline MFNet sometimes focuses on irrelevant regions which bring in redundant temporal information. For example, given the input frames for action ``Wave" as demonstrated in Figure~\ref{visual:wave}, MFNet concentrates not only on the waving hand but also on the face of another actor which is clearly irrelevant to the action ``Wave". In comparison, our MF-KPSEM accurately focuses only on the moving hand of the actress, which are the key regions that imply the ground-truth action.

\paragraph{Visualizing Extracted Key Points and Their Shift}
To further investigate the behaviour of our proposed MF-KPSEM, we visualize six examples over the extracted key points and their corresponding shift as shown in Figure~\ref{detail:visualize-sup}. Our proposed MF-KPSEM could locate key points and their shifts for each action, such as the elbows and feet of the actor for ``Climbing" in Figure~\ref{detail:visualize-1}. The shift marked in arrows match with the actual moves of the actor for the action, which justify the use of key point shifts for extracting temporal features. In addition, in Figure~\ref{detail:visualize-3}, the key point shifts of red crosses indicate that the actor is moving back and forth. These key point shifts describe the characteristic of action sit-up and distinguish it from other similar actions such as rolling forward or lying down. Additionally, in Figure~\ref{detail:visualize-4}, the key point shifts of red and white crosses indicate that the foot shift intensively from the bottom to the top of the video, which is the characteristic of the high kick. It shows that the key point shifts can be applied as the temporal information to differentiate actions from actions.      

The MF-KPSEM is also shown to be able to capture corresponding key points and thus computing their shifts even when key points would alter their location dramatically across regions. Among the samples, Figure~\ref{detail:visualize-1}, Figure~\ref{detail:visualize-3} and Figure~\ref{detail:visualize-5} exhibit the cases where key points remain in the same region across frames. Whereas Figure~\ref{detail:visualize-2}, Figure~\ref{detail:visualize-4} and Figure~\ref{detail:visualize-6} show the possibility that key points could move to another region in the next frame. Take Figure~\ref{detail:visualize-3} and Figure~\ref{detail:visualize-4} as examples for these two scenarios. Key points of ``Sit-up" in Figure~\ref{detail:visualize-3} shift within the same regions while the ones of ``High kick" in Figure~\ref{detail:visualize-4} may shift to a different region in a certain frame. This justifies the need for finding corresponding key points across the frame before extracting the key point shifts in $AReSE$ as mentioned in Section \ref{KPSCompute}.

\section{Conclusion and Future Works}
In this work, we propose a novel method for extracting the temporal features of a video effectively. The new $KPSEM$ exploits key point shifts for temporal feature extraction without additional key point annotation. The overall temporal features encode the key point shifts through linear embedding. Our method obtains state-of-the-art result on Mini-Kinetics when instantiating MFNet, with low additional computational cost compare to other temporal feature extraction methods. We further justify the design and the robustness of our $KPSEM$ module through detailed ablation experiments.

In the future, improvements on explicit key point selection without key point annotation could be further explored. Current key point selection method is effective with trivial computational cost. However, in some cases, it may extract key points irrelevant to the actors or objects of the action. More comprehensive key point selection methods can be developed to further improve extracted temporal features without key point annotation.

\bibliography{main_v1}

\end{document}